\documentclass{IEEEoj}
\usepackage{cite}
\usepackage{amsmath,amssymb,amsfonts}
\usepackage{algorithmic,tikz,pgfplots}
\usepackage{graphicx,color}
\usepackage{bm}

\usepackage[caption=false,font=footnotesize]{subfig}

\usetikzlibrary{positioning,shapes,arrows,arrows.meta,fit,backgrounds,calc}
\usepackage[utf8]{inputenc}
\usepackage{makecell}
\usepackage{multirow}
\usepackage{adjustbox}
\usepackage{textcomp}
\usepackage{booktabs,relsize}
\usepackage{anyfontsize}

\def\BibTeX{{\rm B\kern-.05em{\sc i\kern-.025em b}\kern-.08em
    T\kern-.1667em\lower.7ex\hbox{E}\kern-.125emX}}
\AtBeginDocument{\definecolor{ojcolor}{cmyk}{0.93,0.59,0.15,0.02}}
\def\OJlogo{\vspace{-14pt}\includegraphics[height=28pt]{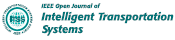}}
\begin{document}
\receiveddate{XX Month, XXXX}
\reviseddate{XX Month, XXXX}
\accepteddate{XX Month, XXXX}
\publisheddate{XX Month, XXXX}
\currentdate{XX Month, XXXX}
\doiinfo{OJITS.2022.1234567}

\title{An Efficient \linebreak Semantic Segmentation Transformer \linebreak for In-Car or Distributed Applications}

\author{DANISH NAZIR\textsuperscript{1,2}, GOWTHAM SAI INTI\textsuperscript{2}, TIMO BARTELS\textsuperscript{1}, JAN PIEWEK\textsuperscript{2}, THORSTEN BAGDONAT\textsuperscript{2}, and TIM FINGSCHEIDT\textsuperscript{1} (Fellow, IEEE)}

\affil{Technische Universität Braunschweig, Braunschweig, Germany}
\affil{Group Innovation, Volkswagen AG, Wolfsburg, Germany}

\corresp{CORRESPONDING AUTHOR: D. NAZIR (e-mail: danish.nazir@volkswagen.de).}

\markboth{Preparation of Papers for IEEE OPEN JOURNALS}{Author \textit{et al.}}

\begin{abstract}

Modern automotive systems leverage deep neural networks (DNNs) for semantic segmentation and operate in two key application areas: (1) In-car, where the DNN solely operates in the vehicle without strict constraints on the data rate. (2) Distributed, where one DNN part operates in the vehicle and the other part typically on a large-scale cloud platform with a particular constraint on transmission bitrate efficiency. Typically, both applications share an image and source encoder, while each uses distinct (joint) source and task decoders. Prior work utilized convolutional neural networks for joint source and task decoding but did not investigate transformer-based alternatives such as \texttt{SegDeformer}, which offer superior performance at the cost of higher computational complexity. In this work, we propose joint feature and task decoding for \texttt{SegDeformer}, thereby enabling \textit{lower computational complexity in both in-car and distributed applications, despite \texttt{SegDeformer}'s  computational demands}. This improves scalability in the cloud while reducing in-car computational complexity. For the in-car application, we increased the frames per second (fps) by up to a factor of $11.7$ ($1.4$ fps to $16.5$ fps) on Cityscapes and by up to a factor of $3.5$ ($43.3$ fps to $154.3$ fps) on ADE20K, while being on-par w.r.t.\ the mean intersection over union (mIoU) of the transformer-based baseline that doesn't compress by a source codec. For the distributed application, we achieve state-of-the-art (SOTA) over a wide range of bitrates on the mIoU metric, while using only $0.14$\% ($0.04$\%) of cloud DNN parameters used in previous SOTA, reported on ADE20K (Cityscapes).

\end{abstract}



\maketitle

\section{INTRODUCTION}
\renewcommand{\thesection}{\Roman{section}}

\renewcommand{\thesubsection}{\thesection.\Alph{subsection}}
\IEEEPARstart{D}{eep} neural networks (DNNs) have achieved state-of-the-art (SOTA) performance on many dense machine perception tasks. Particularly in semantic segmentation \cite{contextualinfo,contextualinfo1}, where the objective is per-pixel classification, transformer-based methods—a new class of DNNs \cite{Segformer,segdeformer}—dominate public benchmarks such as ADE20K \cite{ADE20K} and Cityscapes \cite{cityscapes}. These methods are widely adopted across many applications, including the automotive sector, where they can be deployed as an in-car application through camera-enabled edge devices. However, the computational complexity of the self attention mechanism in transformers  \cite{vaswaniattention}, scaling quadratically with the number of image tokens, together with the limited power and computational capabilities of such devices, makes it challenging to deploy transformers in an in-car application. 

To address this issue, $\texttt{SegFormer}$ \cite{Segformer} replaces standard attention with a more efficient variant and proposes a lightweight multi-layer perceptron (MLP)-based decoder. Although the suggested improvements result in lower computational complexity, the lightweight decoder struggles to capture fine-grained details in complex scenes \cite{segdeformer}, leading to a drop in the mIoU performance. To overcome this limitation, \texttt{SegDeformer} \cite{segdeformer} introduces a powerful transformer-based decoder to capture local and global contexts through hierarchical attention and learnable global class tokens. However, it incurs high computational complexity, especially on high-resolution images, which limits its suitability for real-time in-car application. In this work, we focus on improving the efficiency of \texttt{SegDeformer} while preserving its strong performance for the in-car application.

To overcome the computational and power limitations of the edge devices, recent works \cite{matsubara2021neural,matsubara2019distilled,matsubara2022split,ahuja2023neural,nazirjd} propose a distributed application, which separates the execution of DNNs into two parts, where one part of the DNN is executed on the edge device and the other on a large-scale cloud platform. While the distributed application alleviates the computational and power constraints on the edge device, it also introduces new challenges such as bitrate efficiency and cloud scalability. Also, the prior works focus only on convolutional neural networks (CNNs), leaving the transformer-based state-of-the-art semantic segmentation methods largely unexplored in distributed applications. Further, distributed applications demand a lightweight decoding setup on the cloud for both features and downstream tasks, which is challenging especially with transformer-based methods.

In this article, we make three contributions. First, building upon the SOTA transformer-based task decoder \texttt{SegDeformer}, we propose to perform joint feature and task decoding, which results in an efficient and highly performant cloud DNN, enabling large-scale distributed semantic segmentation without incurring extensive computational load per channel. Second, we extend joint feature and task decoding for the in-car application by introducing a novel \textsf{in-car joint decoding (JD)} method for the \texttt{SegDeformer}, significantly improving its suitability for real-time in-car application. Third, we show that our proposed \textsf{in-car JD} increases the frames per second from $1.4$ fps to $16.5$ fps (factor of $11.7$) on Cityscapes and from $43.3$ fps to $154.3$ fps (factor of $3.5$) on ADE20K, all while maintaining comparable mIoU to the transformer-based \textsf{no compression baseline}. Further, for the distributed application, we set a new distributed semantic segmentation SOTA benchmark over a wide range of bitrates on the mIoU metric, while using only $0.14$\% ($0.04$\%) of the cloud DNN parameters used in previous SOTA, quantified from images of the ADE20K (Cityscapes) datasets.

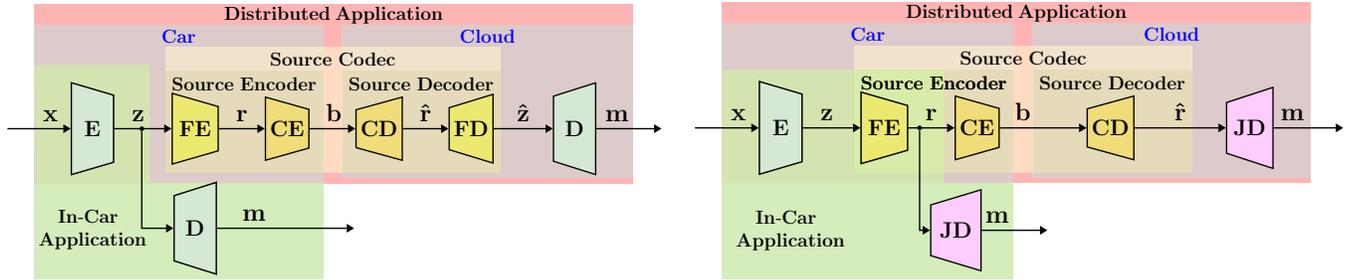
\begin{figure*}[t!]
  \subfloat[{ \raggedright \textbf{Baseline approaches} \textsf{in-car baseline}~\cite{segdeformer} (green) and \textsf{distributed baseline}~\cite{ahuja2023neural} (red).\label{fig:baseline}}]{
    \begin{minipage}[b]{0.48\textwidth}
    \vspace{-0.5em}
        \hspace{-1.5cm}
      \resizebox{1.18\linewidth}{!}{




 \usetikzlibrary{shapes.geometric, arrows.meta, calc,backgrounds,fit}

\definecolor{encoder_decoder}{RGB}{213, 232, 212}
\definecolor{rectangle}{RGB}{255, 245, 204}
\definecolor{se}{RGB}{227, 217, 182}
\definecolor{bottleneck_color}{RGB}{237, 233, 112}
\definecolor{_decoder}{RGB}{255, 204, 255} 
\definecolor{bg}{RGB}{210, 210, 210}
\definecolor{quant_color}{RGB}{232, 123, 142}
\definecolor{ce_color}{RGB}{240, 220, 120}
\definecolor{in_car}{RGB}{204, 255, 153}
\definecolor{light_red}{RGB}{255, 105, 105}

 \tikzstyle{arrow} = [-Triangle, line width=1pt]
 \tikzstyle{label} = [text width= 0.2cm, align=center]
 \tikzstyle{waypoint}=[fill,circle,minimum size=3.5pt,inner sep=0pt]
 \tikzstyle{encoder} = [trapezium,
    trapezium angle=55,
    trapezium stretches=true,
    minimum width=2.4cm,
    minimum height=1.1cm,
    trapezium right angle=80,
    trapezium left angle=80,
    line width=1pt,
    shape border rotate=270,
    text centered,
    draw=black]

 \tikzstyle{bottleneck_encoder} = [trapezium,
    trapezium angle=55,
    trapezium stretches=true,
    minimum width=1.85cm,
    minimum height=1.2cm,
    trapezium right angle=80,
    trapezium left angle=80,
    line width=1pt,
    shape border rotate=270,
    text centered,
    draw=black]
 
 \tikzstyle{compression_encoder} = [trapezium,
    trapezium angle=55,
    trapezium stretches=true,
    minimum width=1.65cm,
    minimum height=1.1cm,
    trapezium right angle=80,
    trapezium left angle=80,
    line width=1pt,
    shape border rotate=270,
    text centered,
    draw=black]

 \tikzstyle{compression_decoder} = [trapezium,
    trapezium angle=55,
    trapezium stretches=true,
    minimum width=1.65cm,
    minimum height=1.1cm,
    trapezium right angle=80,
    trapezium left angle=80,
    line width=1pt,
    shape border rotate=90,
    text centered,
    draw=black]
 \tikzstyle{bottleneck_decoder} = [trapezium,
    trapezium angle=55,
    trapezium stretches=true,
    minimum width=1.8cm,
    minimum height=1.1cm,
    trapezium right angle=80,
    trapezium left angle=80,
    line width=1pt,
    shape border rotate=90,
    text centered,
    draw=black]

  \tikzstyle{decoder} = [trapezium,
    trapezium angle=55,
    trapezium stretches=true,
    minimum width=2.4cm,
    minimum height=1.1cm,
    trapezium right angle=80,
    trapezium left angle=80,
    line width=1pt,
    shape border rotate=90,
    text centered,
    draw=black]

\begin{tikzpicture}[node distance = 1.5cm,font=\fontsize{16.5}{18.5}\selectfont]

\node (true_bg) [draw, rectangle, minimum width=15.6cm, minimum height=6.6cm,  line width=1pt,draw=none, fill=none,yshift=0.cm] at (0,0) {};

\fill[color=light_red, opacity=0.5, line width=1.5pt]
     ([yshift=0.54cm,xshift=0cm]true_bg.north west) rectangle ([xshift=-0.01cm,yshift=2.5cm]true_bg.south east);

\node (background) [draw, rectangle, minimum width=7.5cm, minimum height=6.6cm,  line width=1pt,draw=none, fill=none,yshift=0.0cm,xshift=3.77cm]  at (true_bg.west) {};

\fill[color=bg, opacity=0.75, line width=1.5pt] ([yshift=0cm,xshift=0cm]background.north west) rectangle ([xshift=0cm,yshift=0cm]background.south east);

\node (distributed) [draw, rectangle, minimum width=7.55cm, minimum height=3.96cm,
  line width=1pt, draw=none, fill=none, anchor=north west] 
  at ([xshift=0.5cm]background.north east) {};

\fill[color=bg, opacity=0.75, line width=1.5pt] ([yshift=0cm,xshift=0cm]distributed.north west) rectangle ([xshift=0cm,yshift=0cm]distributed.south east);

\path (background.west); \pgfgetlastxy{\xLeft}{\yBase};
\path (distributed.east); \pgfgetlastxy{\xRight}{\yBase};

\node (cloud_txt) [font=\Large,  label, text width=4cm, above of=distributed, align=center ,yshift=0.2cm, text=blue ] {$\mathbf{Cloud}$};

\node (car_txt) [font=\Large,  label, text width=4cm, above of=background, align=center ,yshift=1.52cm, text=blue ] {$\mathbf{Car}$};

\node[font=\Large, text=black, align=center] at ($(\xLeft, \yBase + 1.2cm)!.5!(\xRight, \yBase + 2.1cm) + (0,0.6cm)$)   {$\mathbf{Distributed \ Application}$};

\node (source_codec) [draw, rectangle, minimum width=8.7cm, minimum height=3.25cm,  
  line width=1pt, draw=none, fill=none, anchor=north, yshift=-0.59cm] 
  at ($(background.north east)!0.51!(distributed.north west)$) {};

\fill[color=rectangle, opacity=0.6, line width=1.5pt] ([yshift=0cm,xshift=0cm]source_codec.north west) rectangle ([xshift=0cm,yshift=0cm]source_codec.south east);

\node[font=\Large,  label, text width=4cm, above of=source_codec, align=center ,yshift=-0.17cm ] {$\mathbf{Source \ Codec}$};
 
\node (source_encoder)[draw, rectangle, minimum width=4.08cm, minimum height=2.62cm,  line width=1pt,draw=none, fill=none, 
right=0.cm of source_codec.west,xshift=0cm,yshift=-0.315cm]  {};
 
\fill[color=se, opacity=0.85, line width=1.5pt] ([yshift=0cm,xshift=0cm]source_encoder.north west) rectangle ([xshift=0cm,yshift=0cm]source_encoder.south east);

\node[font=\Large,  label, text width=4cm, above of=source_encoder, align=center ,yshift=-0.5cm ] {$\mathbf{Source \ Encoder}$};
 
\node (source_decoder)[draw, rectangle, minimum width=4.08cm, minimum height=2.62cm,  line width=1pt,draw=none, fill=none,yshift=-0.315cm, xshift=0cm,right=-4.12cm of source_codec.east] {};

\fill[color=se, opacity=0.85, line width=1.5pt] ([yshift=0cm,xshift=0cm]source_decoder.north west) rectangle ([xshift=0cm,yshift=0cm]source_decoder.south east);

\node[font=\Large,  label, text width=4cm, above of=source_decoder, align=center ,yshift=-0.5cm ] {$\mathbf{Source \ Decoder}$};
 
\node (encoder_1) [encoder , left of=background, xshift=-0.75cm,yshift=0.6cm, fill=encoder_decoder] { $\mathbf{E}$} ;

\node (be_encoder) [bottleneck_encoder , right of=encoder_1,xshift=1.18cm, fill=bottleneck_color] { $\mathbf{FE}$} ;

\node (ce_encoder) [compression_encoder,fill=ce_color, right of=be_encoder,xshift=0.9cm] { $\mathbf{CE}$} ;

\node (ce_decoder) [compression_decoder , right of=ce_encoder,xshift=0.9cm, fill=ce_color] { $\mathbf{CD}$} ;

\node (be_decoder) [bottleneck_decoder, right of=ce_decoder,xshift=0.9cm, fill=bottleneck_color] { $\mathbf{FD}$} ;

\node (decoder_1) [decoder,right of=be_decoder, xshift=1.2cm, fill=encoder_decoder] { $\mathbf{D}$} ;
\node (decoder_2) [decoder,xshift=0.cm,yshift=-1.1cm ,below of=be_encoder, fill=encoder_decoder] { $\mathbf{D}$} ;

\node (incar_bg) [draw, rectangle, minimum width=6cm, minimum height=5.45cm,  line width=1pt,draw=none, fill=none,yshift=-3.9cm,xshift=-0.75cm] at (background.north) {};

\node (fitbox) [draw=none, fit=(incar_bg), inner sep=0pt,yshift=0cm]  {};

\fill[color=in_car, opacity=0.5]
  ([xshift=0.01cm]fitbox.north west) --                                 
  ([xshift=0.03cm,yshift=0cm]fitbox.south west) --                                 
  ([xshift=1.5cm,yshift=0cm]fitbox.south east) --                                  
  ([xshift=1.5cm,yshift=-0.22cm]fitbox.east) --                         
  ([xshift=0.0cm,yshift=-0.22cm]fitbox.center) --                       
  ([xshift=0.0cm,yshift=-0.22]fitbox.center) --                           
  ([xshift=0.cm,yshift=0.1cm]fitbox.north) --                          
  ([xshift=0.01cm,yshift=0.1cm]fitbox.north west) --                    
  cycle;

\node (encoder_1) [encoder , left of=background, xshift=-0.75cm,yshift=0.6cm, fill=encoder_decoder] { $\mathbf{E}$} ;
\node (decoder_2) [decoder,xshift=0.cm,yshift=-1.1cm ,below of=be_encoder, fill=encoder_decoder] { $\mathbf{D}$} ;
\node (in_car_text) [font=\Large,   text width=4cm, above of=fitbox, xshift=-0.4cm,yshift=-2.6cm ] {$\mathbf{In\text{-}Car}$};
\node  [font=\Large,   text width=4cm, below of=in_car_text, xshift=-0.5cm,yshift=0.85cm ] {$\mathbf{Application}$};

\node (way)[waypoint] at ($(encoder_1)!0.48!(be_encoder)$) {};

\draw[arrow] 
  ([xshift=-1.65cm, yshift=0cm]encoder_1.west) 
  to node[midway, xshift=0.3cm, yshift=0.35cm, font=\fontsize{18.5}{20.25}\selectfont] 
  {$\mathbf{x}$} 
  (encoder_1);
\draw [arrow] (encoder_1) to node[midway, xshift=-0.12cm,yshift=0.35cm,font=\fontsize{18.5}{20.5}\selectfont] {$\mathbf{z}$} (be_encoder);

\draw[arrow] (be_encoder) to node[midway, yshift=0.35cm,font=\fontsize{18.5}{20.5}\selectfont] {$\mathbf{r}$} (ce_encoder);

\draw[arrow] (ce_encoder) to node[midway, xshift=0.04cm,yshift=0.42cm,font=\fontsize{18.5}{20.5}\selectfont] {$\mathbf{b}$} (ce_decoder);

\draw[arrow] (ce_decoder) to node[midway, yshift=0.42cm,font=\fontsize{18.5}{20.5}\selectfont] {$\mathbf{\hat{r}}$} (be_decoder);

\draw[arrow] (be_decoder) to node[midway, yshift=0.42cm,font=\fontsize{18.5}{20.5}\selectfont] {$\mathbf{\hat{z}}$} (decoder_1);
\draw[arrow] 
  (decoder_1) 
  to node[midway, xshift=-0.31cm, yshift=0.35cm, font=\fontsize{18.5}{20.5}\selectfont] 
  {$\mathbf{m}$} 
  ([xshift=1.7cm]decoder_1.east);

\draw [arrow] 
  (decoder_2) 
  to node[midway, xshift=-0.82cm, yshift=0.35cm, font=\fontsize{18.5}{20.5}\selectfont] 
  {$\mathbf{m}$} 
  ([xshift=3.58cm]decoder_2.east);

\draw[arrow] (way) -- ++(0,-2.6) to node[midway, xshift=0.05cm,yshift=0.35cm] {} (decoder_2) ;

\end{tikzpicture}

}
    \end{minipage}
  }\hfill
  \subfloat[{  \textbf{Proposed approach} for transformer-based \textsf{in-car joint decoding (JD)} (green) and/or \textsf{distributed joint decoding (JD) (red)}.\label{fig:ours_high_level}}]{
    \begin{minipage}[b]{0.48\textwidth}
    \vspace{-0.5em}
    \hspace{-1.4cm}
      \resizebox{1.18\linewidth}{!}{



\usetikzlibrary{shapes.geometric, arrows.meta, calc,backgrounds}
\usetikzlibrary{fit}
\definecolor{encoder_decoder}{RGB}{213, 232, 212}
\definecolor{rectangle}{RGB}{255, 245, 204}
\definecolor{se}{RGB}{227, 217, 182}
\definecolor{bottleneck_color}{RGB}{237, 233, 112}
\definecolor{_decoder}{RGB}{255, 204, 255} 
\definecolor{bg}{RGB}{210, 210, 210}
\definecolor{quant_color}{RGB}{232, 123, 142}
\definecolor{ce_color}{RGB}{240, 220, 120}
\definecolor{in_car}{RGB}{204, 255, 153}

\definecolor{light_red}{RGB}{255, 105, 105}

 \tikzstyle{arrow} = [-Triangle, line width=1pt]
 \tikzstyle{label} = [text width= 0.2cm, align=center]
 \tikzstyle{waypoint}=[fill,circle,minimum size=3.5pt,inner sep=0pt]
 \tikzstyle{encoder} = [trapezium,
    trapezium angle=55,
    trapezium stretches=true,
    minimum width=2.4cm,
    minimum height=1.1cm,
    trapezium right angle=80,
    trapezium left angle=80,
    line width=1pt,
    shape border rotate=270,
    text centered,
    draw=black]

 \tikzstyle{bottleneck_encoder} = [trapezium,
    trapezium angle=55,
    trapezium stretches=true,
    minimum width=1.85cm,
    minimum height=1.2cm,
    trapezium right angle=80,
    trapezium left angle=80,
    line width=1pt,
    shape border rotate=270,
    text centered,
    draw=black]
  \tikzstyle{compression_decoder} = [trapezium,
    trapezium angle=55,
    trapezium stretches=true,
    minimum width=1.65cm,
    minimum height=1.1cm,
    trapezium right angle=80,
    trapezium left angle=80,
    line width=1pt,
    shape border rotate=90,
    text centered,
    draw=black] 
 \tikzstyle{compression_encoder} = [trapezium,
    trapezium angle=55,
    trapezium stretches=true,
    minimum width=1.65cm,
    minimum height=1.1cm,
    trapezium right angle=80,
    trapezium left angle=80,
    line width=1pt,
    shape border rotate=270,
    text centered,
    draw=black]

 \tikzstyle{bottleneck_decoder} = [trapezium,
    trapezium angle=55,
    trapezium stretches=true,
    minimum width=1.85cm,
    minimum height=1.2cm,
    trapezium right angle=80,
    trapezium left angle=80,
    line width=1pt,
    shape border rotate=90,
    text centered,
    draw=black]

  \tikzstyle{decoder} = [trapezium,
    trapezium angle=55,
    trapezium stretches=true,
    minimum width=1.9cm,
    minimum height=1.1cm,
    trapezium right angle=80,
    trapezium left angle=80,
    line width=1pt,
    shape border rotate=90,
    text centered,
    draw=black]    

\begin{tikzpicture}[node distance = 1.5cm,font=\fontsize{16.5}{18.5}\selectfont]

\node (true_bg) [draw, rectangle, minimum width=15.6cm, minimum height=6.6cm,  line width=1pt,draw=none, fill=none,yshift=0.cm] at (0,0) {};

\fill[color=light_red, opacity=0.5, line width=1.5pt]
     ([yshift=0.54cm,xshift=0cm]true_bg.north west) rectangle ([xshift=-0.415cm,yshift=2.5cm]true_bg.south east);

\node (background) [draw, rectangle, minimum width=7.5cm, minimum height=6.6cm,  line width=1pt,draw=none, fill=none,yshift=0.0cm,xshift=3.77cm]  at (true_bg.west) {};

\fill[color=bg, opacity=0.75, line width=1.5pt] ([yshift=0cm,xshift=0cm]background.north west) rectangle ([xshift=0cm,yshift=0cm]background.south east);

\node (distributed) [draw, rectangle, minimum width=7.15cm, minimum height=3.96cm,
  line width=1pt, draw=none, fill=none, anchor=north west] 
  at ([xshift=0.5cm]background.north east) {};

\fill[color=bg, opacity=0.75, line width=1.5pt] ([yshift=0cm,xshift=0cm]distributed.north west) rectangle ([xshift=0cm,yshift=0cm]distributed.south east);

\path (background.west); \pgfgetlastxy{\xLeft}{\yBase};
\path (distributed.east); \pgfgetlastxy{\xRight}{\yBase};

\node (cloud_txt) [font=\Large,  label, text width=4cm, above of=distributed, align=center ,yshift=0.2cm, text=blue ] {$\mathbf{Cloud}$};
 
\node (car_txt) [font=\Large,  label, text width=4cm, above of=background, align=center ,yshift=1.52cm, text=blue ] {$\mathbf{Car}$};
\node[font=\Large, text=black, align=center] at ($(\xLeft, \yBase + 1.2cm)!.5!(\xRight, \yBase + 2.1cm) + (0,0.6cm)$)   {$\mathbf{Distributed \ Application}$};

\node (source_codec) [draw, rectangle, minimum width=8.7cm, minimum height=3.25cm,  
  line width=1pt, draw=none, fill=none, anchor=north, yshift=-0.59cm] 
  at ($(background.north east)!0.51!(distributed.north west)$) {};

\fill[color=rectangle, opacity=0.6, line width=1.5pt] ([yshift=0cm,xshift=0cm]source_codec.north west) rectangle ([xshift=0cm,yshift=0cm]source_codec.south east);

\node[font=\Large,  label, text width=4cm, above of=source_codec, align=center ,yshift=-0.17cm ] {$\mathbf{Source \ Codec}$};
 
\node (source_encoder)[draw, rectangle, minimum width=4.08cm, minimum height=2.62cm,  line width=1pt,draw=none, fill=none, 
right=0.cm of source_codec.west,xshift=0cm,yshift=-0.315cm]  {};

\fill[color=se, opacity=0.85, line width=1.5pt] ([yshift=0cm,xshift=0cm]source_encoder.north west) rectangle ([xshift=0cm,yshift=0cm]source_encoder.south east);

\node (source_decoder)[draw, rectangle, minimum width=4.08cm, minimum height=2.62cm,  line width=1pt,draw=none, fill=none,yshift=-0.315cm, xshift=0cm,right=-4.12cm of source_codec.east] {};

\fill[color=se, opacity=0.85, line width=1.5pt] ([yshift=0cm,xshift=0cm]source_decoder.north west) rectangle ([xshift=0cm,yshift=0cm]source_decoder.south east);

\node[font=\Large,  label, text width=4cm, above of=source_decoder, align=center ,yshift=-0.5cm ] {$\mathbf{Source \ Decoder}$};

\node[font=\Large,  label, text width=4cm, above of=source_encoder, align=center ,yshift=-0.5cm ] {$\mathbf{Source \ Encoder}$};

\node (encoder_1) [encoder , left of=background, xshift=-0.75cm,yshift=0.6cm, fill=encoder_decoder] { $\mathbf{E}$} ;

\node (be_encoder) [bottleneck_encoder , right of=encoder_1,xshift=1.18cm, fill=bottleneck_color] { $\mathbf{FE}$} ;
\node (ce_encoder) [compression_encoder,fill=ce_color, right of=be_encoder,xshift=0.9cm] { $\mathbf{CE}$} ;
\node (ce_decoder) [compression_decoder , right of=ce_encoder,xshift=1.95cm, fill=ce_color] { $\mathbf{CD}$} ;

\node (incar_bg) [draw, rectangle, minimum width=5.7cm, minimum height=5.35cm,  line width=1pt,draw=none, fill=none,yshift=-3.92cm,xshift=-0.9cm] at (background.north) {};

\node (fitbox) [draw=none, fit=(incar_bg), inner sep=0pt,yshift=0.0cm,xshift=0cm]  {};

 \fill[color=in_car, opacity=0.5]
  ($ (fitbox.north west) + (0.0cm, 0.0cm) $) --                          
  ($ (fitbox.south west) + (0.0cm, 0.0cm) $) --                            
  ($ (fitbox.south east) + (1.8cm, 0.0cm) $) --                           
  ($ (fitbox.east) + (1.8cm, -0.22cm) $) --                               
  ($ (fitbox.center) + (0.0cm, -0.22cm) $) --                             
  ($ (fitbox.center) + (2.87cm, -0.22cm) $) --                               
  ($ (fitbox.north) + (2.87cm, 0.cm) $) --                                
  ($ (fitbox.north west) + (-0.02cm, 0.cm) $) --                          
  cycle;

\node (in_car_text) [font=\Large,   text width=4cm, above of=fitbox, xshift=-0.cm,yshift=-2.6cm ] {$\mathbf{In\text{-}Car}$};
\node  [font=\Large,   text width=4cm, below of=in_car_text, xshift=-0.5cm,yshift=0.85cm ] {$\mathbf{Application}$};

\node[font=\Large,  label, text width=4cm, above of=source_encoder, align=center ,yshift=-0.5cm ] {$\mathbf{Source \ Encoder}$};

\node (encoder_1) [encoder , left of=background, xshift=-0.75cm,yshift=0.6cm, fill=encoder_decoder] { $\mathbf{E}$} ;
\node (be_encoder) [bottleneck_encoder , right of=encoder_1,xshift=1.18cm, fill=bottleneck_color] { $\mathbf{FE}$} ;

\node (way)[waypoint] at ($(be_encoder)!0.38!(ce_encoder)$) {};
 
\node (decoder_1) [decoder,right of=ce_decoder, xshift=2.1cm, fill=encoder_decoder,minimum width=2.1cm, minimum height=1.2cm,fill=_decoder] { $\mathbf{JD}$} ;
\node (decoder_2) [decoder,xshift=-0.55cm,yshift=-1.15cm ,below of=ce_encoder, fill=_decoder,minimum width=2.1cm, minimum height=1.3cm] { $\mathbf{JD}$} ;

\draw[arrow] 
  ([xshift=-1.65cm, yshift=0cm]encoder_1.west) 
  to node[midway, xshift=0.3cm, yshift=0.35cm, font=\fontsize{18.5}{20.25}\selectfont] 
  {$\mathbf{x}$} 
  (encoder_1);
  
\draw [arrow] (encoder_1) to node[midway, xshift=-0.12cm,yshift=0.35cm,font=\fontsize{18.5}{20.5}\selectfont] {$\mathbf{z}$} (be_encoder);

\draw[arrow] (be_encoder) to node[midway, yshift=0.35cm,font=\fontsize{18.5}{20.5}\selectfont] {$\mathbf{r}$} (ce_encoder);

\draw[arrow] (ce_encoder) to node[midway, xshift=-0.52cm,yshift=0.42cm,font=\fontsize{18.5}{20.5}\selectfont] {$\mathbf{b}$} (ce_decoder);

\draw[arrow] (ce_decoder) to node[midway, yshift=0.42cm,font=\fontsize{18.5}{20.5}\selectfont] {$\mathbf{\hat{r}}$} (decoder_1);

\draw[arrow] (way) -- ++(0,-2.65) to node[midway, xshift=0.05cm,yshift=0.35cm] {} (decoder_2) ;

\draw[arrow] 
  (decoder_1) 
  to node[midway, xshift=-0.41cm, yshift=0.35cm, font=\fontsize{18.5}{20.5}\selectfont] 
  {$\mathbf{m}$} 
  ([xshift=1.8cm]decoder_1.east);

\draw [arrow] 
  (decoder_2) 
  to node[midway, xshift=-0.45cm, yshift=0.35cm, font=\fontsize{18.5}{20.5}\selectfont] 
  {$\mathbf{m}$} 
  ([xshift=1.7cm]decoder_2.east);
\end{tikzpicture}

}
    \end{minipage}
  }
  \caption{  
\textbf{High-level comparison} of \textbf{existing state-of-the-art} vs.\ \textbf{our proposed approaches} in both in-car (green) and distributed (red) semantic segmentation applications. Here, $\mathbf{E}$ and $\mathbf{D}$ are the image encoder and task decoder, respectively. Blocks $\mathbf{FE}$ and $\mathbf{FD}$ represent the feature encoder and decoder, respectively, while $\mathbf{CE}$ and $\mathbf{CD}$ correspond to the compression encoder and decoder. $\mathbf{JD}$ represents the proposed transformer-based joint feature and task decoder. }
  \label{fig:high_level}
\end{figure*}

\section{FROM RELATED WORKS TO OUR PROPOSAL}
\label{sec:related_to_proposal}

In this section, we begin with an overview of semantic segmentation methods, followed by their applications in the automotive domain and extension to the distributed setting.

\subsection{SEMANTIC SEGMENTATION}
Semantic segmentation is a dense prediction task that aims to classify each pixel of an image with a set of predefined categories. The advent of fully convolutional deep neural networks (FCNs) \cite{long2015fully,FCN2,FCN3,badrinarayanan2017segnet,ronneberger2015u} enabled semantic segmentation with reduced computational demands, while achieving state-of-the-art (SOTA) performance on various public benchmarks \cite{cityscapes,pascalvoc,cocodataset,ADE20K}. Subsequent works proposed further improvements in FCNs, such as dilated \cite{dialted1,dialted2} and deformable convolutions \cite{dai2017deformable}, atrous spatial pyramid pooling \cite{deeplabv3,chen2018atrous}, and feature fusion \cite{wang2020deep}. More recently, transformer-based methods \cite{Segformer,segdeformer} have emerged as the new SOTA in the field of semantic segmentation due to their ability to efficiently model long-range dependencies, while maintaining comparable computational complexity to FCNs. To further improve their efficiency, \texttt{SegFormer} \cite{Segformer} and other recent approaches \cite{jain2023semask,shi2022ssformer,yan2024multi,he2023shifted} have introduced lightweight transformer-based encoder architectures, along with multi-layer perceptrons (MLP) and FCN-based decoders. Although these MLP and FCN-based decoders are computationally efficient, they have limited expressive power, which restricts their performance in complex scenes\cite{segdeformer}. To address this limitation, \texttt{SegDeformer} \cite{segdeformer} has introduced a powerful transformer-based decoder for both internal and external context mining, enabling it to effectively learn complex local and global contexts, thereby improving segmentation performance in challenging environments. In this work, we will build upon \texttt{SegDeformer} \cite{segdeformer}, since it overcomes the limited expressive power of MLP and FCN-based decoders.

\subsection{SEMANTIC SEGMENTATION IN AUTOMOTIVE APPLICATIONS}

Semantic segmentation is a fundamental component of the perception stack in autonomous driving \cite{feng2020deep,houben2022inspect,wang2025reliable}, supporting various downstream tasks such as lane detection \cite{lim2017implementation,liu2024lane}, drivable area estimation \cite{park2021drivable,qiao2021drivable}, and road user classification. In automotive applications, it poses additional challenges, including the need for real-time inference, low latency, and adaptability to diverse and dynamic environments \cite{papadeas2021real}. To fulfill these constraints, semantic segmentation deep neural networks (DNNs) are typically deployed as an in-car application, where they are completely executed inside the vehicle using its limited onboard computational resources. However, this setup imposes strict constraints on DNN size and energy consumption. Due to such strict constraints, CNN-based methods \cite{deeplabv3,yu2021bisenet,zhao2017pyramid} are preferred due to their low computational complexity. However, their limited receptive field prevents them to capture long-range dependencies and complex local and global contexts \cite{segdeformer}, which are critical in complex driving scenes. To address this limitation, \texttt{SegDeformer} \cite{segdeformer} employs a transformer-based task decoder. Despite its strong performance, it incurs high computational complexity, making it inefficient for an in-car application. Therefore, inspired by \cite{nazirjd}, we introduce a novel joint feature and task decoding (\textsf{in-car joint decoding (JD)}) to significantly reduce the computational complexity of \texttt{SegDeformer} to enhance its suitability for real-time in-car application.

In Figure \ref{fig:high_level}, we present a high-level comparison between \texttt{SegDeformer} \cite{segdeformer}, which we refer to as \textsf{in-car baseline} (left), and our proposed transformer-based \textsf{in-car JD} (right). Fig.\ \ref{fig:high_level}\subref{fig:baseline} shows that the \textsf{in-car baseline} deploys both an image encoder $\mathbf{E}$ and a task decoder $\mathbf{D}$ in the car to generate a semantic segmentation map $\mathbf{m} \in \mathcal{S}^{H \times W}$ with height $H$, width $W$, classes $\mathcal{S}=\{ 1,2, .., S \}$  and number of classes $S$. Fig.\ \ref{fig:high_level}\subref{fig:ours_high_level} illustrates the proposed \textsf{in-car JD}, which utilizes $\mathbf{E}$ and a low-complex source encoder in the car to compress the bottleneck features $\mathbf{z} \in \mathbb{R}^{F \times \frac{H}{4} \times \frac{W}{4}}$ with the number of kernels $F$, consisting of a feature encoder $\mathbf{FE}$ to generate the latent representation $\mathbf{r} \in \mathbb{R}^{F \times \frac{H}{8} \times \frac{W}{8}}$, which is then passed to the \textit{proposed transformer-based joint feature and task decoder $\mathbf{JD}$ to generate $\mathbf{m}$}. 




\subsection{DISTRIBUTED SEMANTIC SEGMENTATION}

Modern perception systems in autonomous vehicles employ multiple high-resolution cameras, which significantly increase the computational and energy demands of the in-car processing unit. As a result, it may not be feasible to fully execute the semantic segmentation DNN within the car. This challenge has led to the rise of distributed semantic segmentation, which distributes the execution of the semantic segmentation DNN over a client$/$server architecture 
\cite{lohdefink2019low,liu2022improving,torfason2018towards,wang2022learning,liu2022semantic}, as also shown in Figure \ref{fig:high_level}. Although the distributed application mitigates the computational and energy limitations of the in-car application, it introduces two additional constraints: (1) bitrate efficiency for transmitting the bottleneck features $\mathbf{z}$ from the vehicle to the cloud, and (2) scalability of the service without incurring high computational load on the cloud. To enable bitrate efficient communication between the vehicle and the cloud, prior works \cite{ahuja2023neural,nazirjd,matsubara2022bottlefit,matsubara2022sc2,matsubara2019distilled,shao2020bottlenet,matsubara2021neural} have introduced low-complexity source codecs, with the source encoder being deployed on the vehicle and the source decoder on the cloud. The source encoder receives the bottleneck features $\mathbf{z}$ as input and produces a quantized compressed bitstream $\mathbf{b}$, which is transmitted to the cloud. In the cloud, the source decoder reconstructs the bottleneck features $\mathbf{\hat{z}}$, which are then passed to the task decoder. While prior works provide a way to enable bitrate-efficient transmission, they do not address the challenge of improving the scalability of the distributed semantic segmentation service. To improve the scalability of the distributed semantic segmentation, recent work proposes a convolutional neural network (CNN)-based joint source and task decoder \cite{nazirjd}, while achieving SOTA rate-distortion (RD) performance on various distributed semantic segmentation benchmarks \cite{cityscapes,cocodataset}. However, \cite{nazirjd} does not consider SOTA transformer-based task decoders such as \texttt{SegDeformer}. In this work, we propose joint feature and task decoding (\textsf{distributed JD}) for the \texttt{SegDeformer} to increase RD performance, while improving the scalability of the distributed semantic segmentation service. 

In Figure \ref{fig:high_level}, we illustrate on high-level our proposed transformer-based \textsf{distributed joint decoding (JD)} against the \textsf{distributed baseline} \cite{ahuja2023neural}. Fig.\ \ref{fig:high_level}\subref{fig:baseline} shows that the \textsf{distributed baseline} deploys $\mathbf{E}$ along with a low-complex source encoder inside the car to compress the bottleneck features $\mathbf{z} \in \mathbb{R}^{F \times \frac{H}{4} \times \frac{W}{4}}$ with the number of kernels $F$. It consists of a feature encoder $\mathbf{FE}$ and a compression encoder $\mathbf{CE}$. The source encoder outputs a quantized compressed bitstream $\mathbf{b}$, which is transmitted to the cloud. In the cloud, a respective source decoder, consisting of compression decoder $\mathbf{CD}$ and feature decoder $\mathbf{FD}$ is utilized to reconstruct the bottleneck features $\mathbf{\hat{z}} \in \mathbb{R}^{F \times \frac{H}{4} \times \frac{W}{4}}$, which are passed to $\mathbf{D}$ to compute $\mathbf{m}$ on the receiving side. Fig.\ \ref{fig:high_level}\subref{fig:ours_high_level} depicts the proposed \textsf{distributed JD}. The quantized compressed bitstream $\mathbf{b}$ is sent to the respective source decoder deployed in the cloud. In our case, the source task decoder contains only $\mathbf{CD}$ and reconstructs the latent representation $\mathbf{\hat{r}}$, which is then utilized \textit{by the proposed transformer-based joint feature and task decoder $\mathbf{JD}$ to generate $\mathbf{m}$ in a single step}.

In this work, we build upon the current SOTA in distributed semantic segmentation \cite{nazirjd} and propose joint feature and task decoding building upon the SOTA transformer-based task decoder \texttt{SegDeformer}. This further increases both the RD performance and the scalability of the distributed semantic segmentation application. Furthermore, we propose a novel \textsf{in-car JD} scheme for \texttt{SegDeformer} to significantly reduce its computational complexity for the in-car application.

\section{METHOD}

While Section \ref{sec:related_to_proposal} already introduced our proposed approach on high-level, here we will go deeper describing an efficient source codec for bottleneck feature compression, followed by our proposed approach for joint feature and task decoding for \texttt{SegDeformer}.

\subsection{EFFICIENT SOURCE CODEC FOR BOTTLENECK FEATURE COMPRESSION}
\label{sec:efficient_source_codec}

Recent works \cite{matsubara2022sc2,ahuja2023neural,matsubara2021neural} have proposed to employ source codecs for enabling bitrate-efficient communication between the edge devices, such as vehicles, and the cloud. The source codec is used for compressing the bottleneck features $\bf{z=E(x;\bm{\theta}^{\mathrm{E}})}$$\ \in \mathbb{R}^{F \times \frac{H}{4} \times \frac{W}{4}}$ from an image encoder $\bf{E}$ with parameters $\bm{\theta}^{\mathrm{E}}$. Here, $\bf{x} = (\bf{x}_{\textit{i}}) \in \mathbb{I}$$^{C \!\times\! H \!\times\! W}$ is a normalized image of $C=3$ color channels, height $H$ and width $W$, with pixel $\bf{x}_{\textit{i}} \in \mathbb{I}$$^C$, pixel index $i \in \mathcal{I}$, pixel index set $\mathcal{I} = \{ 1,...,H \cdot W \}$ and $\mathbb{I}=[0,1]$. The source codec consists of a bottleneck source encoder and a bottleneck source decoder. The bottleneck source encoder is executed in-car, while the bottleneck source decoder is executed on the cloud.

\begin{figure*}[t!]
    \centering
    \resizebox{0.85\linewidth}{!}{ \usetikzlibrary{shapes.geometric, arrows.meta, calc,backgrounds}

\definecolor{encoder_decoder}{RGB}{213, 232, 212}
\definecolor{rectangle}{RGB}{255, 245, 204}
\definecolor{bottleneck_color}{RGB}{237, 233, 112}
\definecolor{_decoder}{RGB}{255, 204, 255} 
\definecolor{conv}{RGB}{255, 255, 255}
\definecolor{quant_color}{RGB}{232, 123, 142}
\definecolor{ae_color}{RGB}{221, 221, 221}
\definecolor{_rectangle}{RGB}{221, 221, 221}
\definecolor{source_color}{RGB}{200, 201, 164}
\definecolor{hyper_color}{RGB}{255, 219, 77}
\definecolor{ce_color}{RGB}{230, 221, 184}

\tikzstyle{process} = [rectangle,minimum width=5cm, minimum height=1cm, text
centered, draw=black, text=black,line width = 1pt, fill=conv]
 \tikzstyle{arrow} = [-Triangle, line width=1pt]
 \tikzstyle{label} = [text width= 0.2cm, align=center]
 \tikzstyle{waypoint}=[fill,circle,minimum size=3.5pt,inner sep=0pt]
 \tikzstyle{encoder} = [trapezium,
    trapezium angle=55,
    trapezium stretches=true,
    minimum width=7.95cm,
    minimum height=5.2cm,
    trapezium right angle=90,
    trapezium left angle=90,
    shape border rotate=180,
    text centered,
    ]

  \tikzstyle{decoder} = [trapezium,
    trapezium angle=55,
    trapezium stretches=true,
    minimum width=8cm,
    minimum height=4.6cm,
    trapezium right angle=85,
    trapezium left angle=85,
    line width=1.5pt,
    shape border rotate=180,
    text centered,
    draw=black]    

\begin{tikzpicture}[node distance = 1.5cm,font=\LARGE]

\node (source_encoder) [draw, rectangle, minimum width=17cm, minimum height=12cm,  line width=0pt,draw=none, fill=ce_color,yshift=0cm]   {};
\node  (source_decoder) [draw, rectangle, minimum width=8.8cm, minimum height=12cm,  line width=0pt,draw=none, fill=ce_color,yshift=0cm,xshift=12.13cm,  right of=source_encoder] {};

\node (heading) [above of=source_encoder,xshift=0cm,yshift=3.9cm, font=\LARGE]{$\mathbf{Compression \ Encoder \ (CE) }$};
\node (heading2) [above of=source_decoder,xshift=0cm,yshift=3.9cm, font=\LARGE]  {$\mathbf{Compression \ Decoder \ (CD)} $ };

\node (hyperprior_encoder) [encoder ,minimum width=7.95cm,  xshift=-4cm, yshift=-0.35cm,  fill=hyper_color] at (source_encoder) { } ;
\node (heading) [above of=hyperprior_encoder,xshift=-3.25cm,yshift=0.7cm]  {$\mathbf{HE}$};
\node (he_conv1) [process, yshift=1.35cm,minimum width=7.4cm,minimum height=0.9cm] at (hyperprior_encoder) {$\mathrm{Conv(3 \times 3, }F,\rho=2)$};
\node (he_batchnorm) [process,minimum width=7.4cm,yshift=0.4cm, minimum height=0.9cm,below of=he_conv1]  {$\mathrm{BatchNorm + ReLU}$};
\node (he_conv2) [process, yshift=0.4cm,minimum width=7.4cm,minimum height=0.9cm,below of=he_batchnorm]  {$\mathrm{Conv(1 \times 1, }F)$};
\node (he_batchnorm2) [process,minimum width=7.4cm,yshift=0.4cm, minimum height=0.9cm,below of=he_conv2]  {$\mathrm{BatchNorm + ReLU}$};
\node (z) [label, xshift=1.18cm,yshift=-0.7cm,below of=he_batchnorm2,text width=1.89cm,font=\LARGE] {$\mathbf{h} \in \mathcal{H}$};
\node (quant_hyperprior) [draw, rectangle, text=black, minimum width=1.5cm, minimum height=1.5cm,  line width=1.5pt,draw=black, fill=quant_color,xshift=3.cm,yshift=-1.2cm,below of=he_batchnorm2,font=\LARGE]   {$\mathbf{Q^{\mathcal{H}}}$};
\node (ae_hyperprior) [draw, rectangle, text=black, minimum width=1.5cm, minimum height=1.5cm,  line width=1.5pt,draw=black, fill=ae_color,yshift=0cm,xshift=1.5cm,right of=quant_hyperprior,font=\LARGE]   {$\mathbf{AE^{\mathcal{H}}}$};

\node (ad_hyperprior) [draw, rectangle, text=black, minimum width=1.5cm, minimum height=1.5cm,  line width=1.5pt,draw=black, fill=ae_color,yshift=1.14cm,xshift=0.8cm,right of=ae_hyperprior,font=\LARGE]   {$\mathbf{AD^{\mathcal{H}}}$};
\node (way_hyper)[waypoint,yshift=-1.13cm] at ($(ae_hyperprior)!1!(ad_hyperprior)$) {};

\node (hyperprior_decoder) [encoder , yshift=0cm,xshift=6.8cm,  fill=hyper_color, right of=hyperprior_encoder]  { } ;
\node (sigma) [label, xshift=0.5cm,yshift=1.5cm ,above of=hyperprior_decoder,text width=1.39cm,font=\LARGE] {$\boldsymbol{\sigma}$};

\node (quant_latent_orig) [draw, rectangle, text=black, minimum width=1.5cm, minimum height=1.5cm,  line width=1.5pt, draw=black,yshift=2.8cm,xshift=-5.31cm,above of =hyperprior_decoder,fill=quant_color,font=\LARGE]   {$\mathbf{Q^{\mathcal{R}}}$};

\node (ae_latent)  [draw, rectangle, text=black, right of =quant_latent_orig, minimum width=1.5cm, minimum height=1.5cm,  line width=1.5pt,draw=black, fill=ae_color,yshift=0cm,xshift=1.5cm,font=\LARGE]  {$\mathbf{AE^{\mathcal{R}}}$};
\node (quant_latent) [draw, rectangle, text=black, minimum width=1.5cm, minimum height=1.5cm,  line width=1.5pt, draw=none,yshift=0cm,xshift=-6.02cm,left of=ae_latent,font=\LARGE]   {};

\node (way)[waypoint] at ($(quant_latent)!0.338!(quant_latent_orig)$) {};

\node (y) [label, xshift=0.3cm,yshift=0cm,left of=way,text width=1.89cm,font=\LARGE] {$\mathbf{r} \in \mathcal{R}$};
\node (fake_input) [draw, rectangle, text=black, minimum width=1.5cm, minimum height=1.5cm,  line width=1.5pt,draw=none, fill=none,yshift=2.9cm,xshift=0cm,above of=he_conv1,font=\LARGE]   {};
\node [above of=hyperprior_decoder,xshift=-3.2cm,yshift=0.7cm] (heading) {$\mathbf{HD}$};

\node (hd_conv1) [process, yshift=1.35cm,minimum width=7.4cm,minimum height=0.9cm] at (hyperprior_decoder) {$\mathrm{BatchNorm + ReLU}$};
\node (hd_batchnorm) [process,minimum width=7.4cm,yshift=0.4cm, minimum height=0.9cm,below of=hd_conv1]  {$\mathrm{UpConv(3 \times 3, }F,\rho=2)$};
\node (hd_conv2) [process, yshift=0.4cm,minimum width=7.4cm,minimum height=0.9cm,below of=hd_batchnorm]  {$\mathrm{BatchNorm + ReLU}$};
\node (hd_batchnorm2) [process,minimum width=7.4cm,yshift=0.4cm, minimum height=0.9cm,below of=hd_conv2]  {$\mathrm{UpConv(1 \times 1, }F)$};

\node (text) [anchor=east,below of=hd_batchnorm2,yshift=0.7cm,xshift=0.3cm]  {};
\node (label) [right of=ad_hyperprior,xshift = 0.5cm,yshift=0.5cm]  {$\mathbf{\hat{h}}$};

\node (ad_latent_source) [draw, rectangle, text=black, minimum width=1.5cm, minimum height=1.5cm,  line width=1.5pt,draw=black, fill=ae_color,yshift=0cm,xshift=10.13cm,right of=ae_hyperprior,font=\LARGE]   {$\mathbf{AD^{\mathcal{H}}}$};

\node (cd_hyperprior_decoder) [encoder , yshift=0cm,xshift=7.82cm,  fill=hyper_color, right of=hyperprior_decoder]  { } ;
\node [above of=cd_hyperprior_decoder,xshift=-3.2cm,yshift=0.7cm] (heading) {$\mathbf{HD}$};
\node (cd_hd_conv1) [process, yshift=1.35cm,minimum width=7.4cm,minimum height=0.9cm] at (cd_hyperprior_decoder) {$\mathrm{BatchNorm + ReLU}$};
\node (cd_hd_batchnorm) [process,minimum width=7.4cm,yshift=0.4cm, minimum height=0.9cm,below of=cd_hd_conv1]  {$\mathrm{UpConv(3 \times 3, }F,\rho=2)$};
\node (cd_hd_conv2) [process, yshift=0.4cm,minimum width=7.4cm,minimum height=0.9cm,below of=cd_hd_batchnorm]  {$\mathrm{BatchNorm + ReLU}$};
\node (cd_hd_batchnorm2) [process,minimum width=7.4cm,yshift=0.4cm, minimum height=0.9cm,below of=cd_hd_conv2]  {$\mathrm{UpConv(1 \times 1, }F)$};

\node (ad_hyper_source) [draw, rectangle, text=black, minimum width=1.5cm, minimum height=1.5cm,  line width=1.5pt,draw=black, fill=ae_color,yshift=2.8cm,xshift=0cm,above of=cd_hyperprior_decoder,font=\LARGE]   {$\mathbf{AD^{\mathcal{R}}}$};
\node (fake_output) [draw, rectangle, text=black, minimum width=1.5cm, minimum height=1.5cm,  line width=1.5pt,draw=none, fill=none,yshift=0cm,xshift=4.3cm,right of=ad_hyper_source,font=\LARGE]   {};

\draw [arrow] (quant_hyperprior) to node[midway, xshift=0cm,yshift=0.57cm,font=\LARGE] {$\mathbf{\hat{h}}$} (ae_hyperprior);
\draw[arrow] (he_batchnorm2) -- ++(0,-0.48)  -- ++(0,-2.21)  --  (quant_hyperprior.west);
\draw[arrow] (ae_hyperprior) -- ++(2.3,0)  -- ++(0,0)  --  (ad_hyperprior.south);
\draw [arrow] (way) to (quant_latent_orig);
\draw [arrow] (fake_input) to (he_conv1);
\draw[arrow] (ad_hyperprior) --   (hd_batchnorm2) ;
\draw (label.west) edge[out=180,in=40,-,line width=1pt] (text);
\draw[arrow] (hd_conv1) -- ++(0,1.6)  -- ++(-2.3,0)  --   (ae_latent.south);
\draw[arrow] (way_hyper) to node[midway, xshift=0.3cm,yshift=0.45cm,font=\LARGE] {$\mathbf{b_{\hat{h}}}$}  (ad_latent_source) ;
\draw [arrow] (ad_latent_source) to  node[midway, above, align=center,xshift=0.4cm,yshift=-0.4cm,font=\LARGE] {$\mathbf{{\hat{h}}}$}    (cd_hd_batchnorm2);
\draw [arrow] (quant_latent_orig) to  node[midway, above, align=center,xshift=0.cm,yshift=0.15cm,font=\LARGE] {$\mathbf{{\hat{r}}}$}    (ae_latent);

\draw [arrow] (ae_latent) to node[midway, xshift=1.05cm,yshift=0.45cm,font=\LARGE] {$\mathbf{b_{\hat{r}}}$} (ad_hyper_source);
\draw [arrow] (cd_hd_conv1)    to  node[midway, above, align=center,xshift=0.4cm,yshift=0cm,font=\LARGE] {$\boldsymbol{\sigma}$}   (ad_hyper_source);
\draw[arrow] (ad_hyper_source) to  node[midway, above, align=center,xshift=1cm,yshift=0.15cm,font=\LARGE] {$\mathbf{{\hat{r}}}$}   (fake_output) ;

\end{tikzpicture}}
  \caption{Hyperprior architecture of \textbf{compression encoder} $\mathbf{CE}$ and \textbf{compression decoder} $\mathbf{CD}$ utilized in Fig.\ \ref{fig:high_level} for the {\normalfont \textsf{distributed baseline}}, {\normalfont \textsf{distributed JD baseline}}, and for the proposed transformer-based {\normalfont \textsf{distributed JD}}, adopted from \cite{nazirjd}.  
  } 
  \label{fig:ce_cd}
\end{figure*}
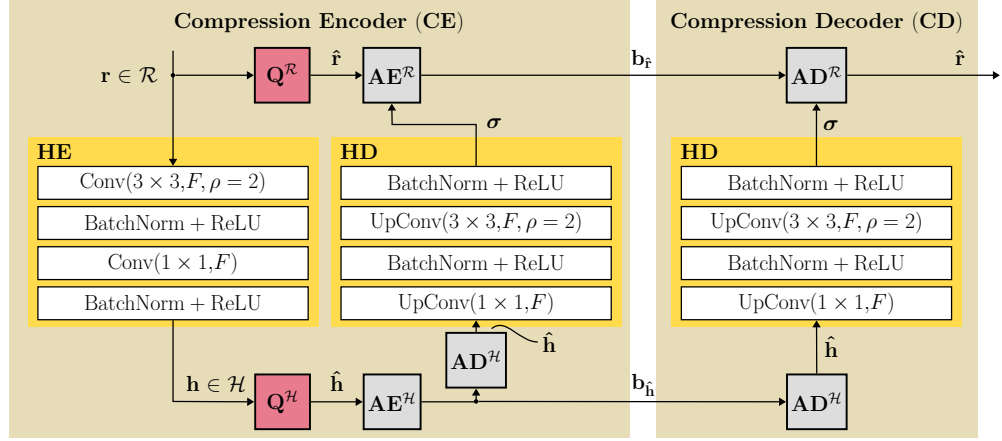

Fig.\ \ref{fig:high_level}\subref{fig:baseline} illustrates the \textsf{distributed baseline} \cite{ahuja2023neural}, where the bottleneck source encoder projects bottleneck features $\bf{z}$ to a latent representation $\bf{r=FE(z;\bm{\theta}^{\mathrm{FE}})}$ $\in \mathbb{R}^{F \times \frac{H}{8} \times \frac{W}{8}}$ with a feature encoder $\mathbf{FE}$ having parameters $\bm{\theta}^{\mathrm{FE}}$, followed by a compression encoder $\bf{CE}$ with parameters $\bm{\theta}^{\mathrm{CE}}$ to generate a latent core bitstream $\bf{b}_{\hat{r}}$ and enhancement bitstream $\bf{b}_{\hat{h}}$, both jointly denoted as $\bf{b=(\bf{b}_{\hat{r}},\bf{b}_{\hat{h}})=CE(r;\bm{\theta}^{\mathrm{CE}})}$. The bottleneck source decoder proceeds by reconstructing the latent representation through a compression decoder $\bf{CD}$ with parameters $\bm{\theta}^{\mathrm{CD}}$ as $\bf{\hat{r}}=\bf{CD}(\bf{b};\bm{\theta}^{\mathrm{CD}})$ $\in \mathbb{R}^{F \times \frac{H}{8} \times \frac{W}{8}}$, followed by bottleneck features reconstruction via a feature decoder $\bf{FD}$ with parameters $\bm{\theta}^{\mathrm{FD}}$ as $\bf{\hat{z}}=\bf{FD}(\bf{\hat{r}};\bm{\theta}^{\mathrm{FD}})$ $\in \mathbb{R}^{F \times \frac{H}{4} \times \frac{W}{4}}$, which are passed to the distributed task decoder $\bf{D}$ with parameters $\bm{\theta}^{\mathrm{D}}$ to generate a semantic segmentation map $\bf{m}=\bf{D(z;\bm{\theta}^{\mathrm{D}})}$. Fig.\ \ref{fig:high_level}\subref{fig:baseline}
 also shows the \textsf{in-car baseline} \cite{segdeformer}, where the bottleneck features $\mathbf{z}$ are directly passed to the in-car task decoder $\bf{D}$ with parameters $\bm{\theta}^{\mathrm{D}}$ resulting in semantic segmentation map $\bf{m}=\bf{D(z;\bm{\theta}^{\mathrm{D}})}$. Note that the network architectures of both in-car and distributed task decoders $\bf{D}$ in Fig.\ \ref{fig:high_level}\subref{fig:baseline} are exactly the same. The detailed network architectures of feature encoder $\mathbf{FE}$, feature decoder $\mathbf{FD}$, compression encoder $\mathbf{CE}$, compression decoder $\mathbf{CD}$, task decoder $\mathbf{D}$, are shown in Figures \ref{fig:fe_fd}, \ref{fig:ce_cd}, \ref{fig:overview_architectures}(a), respectively, all described in the following.

In Figure \ref{fig:fe_fd}, the network architectures of the feature encoder $\mathbf{FE}$ and feature decoder $\mathbf{FD}$ are illustrated. The $\mathbf{FE}$ takes the bottleneck features $\mathbf{z}$ as input and produces a latent representation $\mathbf{r}$. Further, the $\mathbf{FD}$ receives the quantized latent representation $\mathbf{\hat{r}}$ as input and outputs the reconstructed bottleneck features $\mathbf{\hat{z}}$. Both, $\bf{FE}$ and $\bf{FD}$, contain grouped (i.e., depthwise) convolutions denoted by $\mathrm{DWConv(\textit{h} \times \textit{h},\textit{F},\textit{G},\rho=2)}$ and grouped transposed convolutions $\mathrm{DWUpConv(\textit{h} \times \textit{h},\textit{F},\textit{G},\rho=2)}$ \cite{depthwiseseparable}, where $\mathrm{\textit{h} \times \textit{h}}$ is the kernel size, $F$ is the number of kernels, $G$ is the number of groups present in the layer, and stride $\rho=2$. They also contain reg\-ular convolutional and transposed convolutional layers, which are represented by $\mathrm{Conv(\textit{h} \times \textit{h},\textit{F})}$ and $\mathrm{UpConv(\textit{h} \times \textit{h},\textit{F})}$, respectively.

\begin{figure}[t!]
 
  \begin{minipage}[t]{0.49\linewidth}
  \centering
   \hspace{-1cm}
    \resizebox{1.17\linewidth}{!}{\usetikzlibrary{shapes.geometric, arrows.meta, calc,backgrounds}

\definecolor{encoder_decoder}{RGB}{213, 232, 212}
\definecolor{rectangle}{RGB}{255, 245, 204}
\definecolor{bottleneck_color}{RGB}{237, 233, 112}
\definecolor{_decoder}{RGB}{255, 204, 255} 
\definecolor{conv}{RGB}{255, 255, 255}
\definecolor{quant_color}{RGB}{255, 162, 234}
\definecolor{ae_color}{RGB}{221, 221, 221}
\definecolor{_rectangle}{RGB}{221, 221, 221}
\definecolor{source_color}{RGB}{200, 201, 164}
\definecolor{hyper_color}{RGB}{255, 219, 77}

\tikzstyle{process} = [rectangle,minimum width=5cm, text
centered, draw=black, text=black,line width = 1pt, fill=conv]
 \tikzstyle{arrow} = [-Triangle, line width=1pt]
 \tikzstyle{label} = [text width= 0.2cm, align=center]
 \tikzstyle{waypoint}=[fill,circle,minimum size=3.5pt,inner sep=0pt]
 \tikzstyle{encoder} = [trapezium,
    trapezium angle=55,
    trapezium stretches=true,
    minimum width=7.95cm,
    minimum height=5.2cm,
    trapezium right angle=90,
    trapezium left angle=90,
    shape border rotate=180,
    text centered,
    ]

  \tikzstyle{decoder} = [trapezium,
    trapezium angle=55,
    trapezium stretches=true,
    minimum width=8cm,
    minimum height=4.6cm,
    trapezium right angle=85,
    trapezium left angle=85,
    line width=1.5pt,
    shape border rotate=180,
    text centered,
    draw=black]

\begin{tikzpicture}[node distance = 1.5cm,font=\fontsize{24}{26}\selectfont]

\node (background) [draw, rectangle, minimum width=11.6cm, minimum height=5.65cm,  draw=none, fill=bottleneck_color]   {};

\node (conv1) [process,xshift=0cm, yshift=0.42cm,minimum width=10.88cm,minimum height=0.9cm, above of=background,]  {$\mathrm{DWConv(3 \times 3,}F,G,\rho=2)$};

\node at ($(conv1)+(-0.5cm,1.45cm)$) [font=\fontsize{26.5}{28.5}\selectfont] (labels) {$\mathbf{z}$};

\node (batchnorm) [process,minimum width=10.88cm,yshift=0.18cm, minimum height=1.03cm,below of=conv1]  {$\mathrm{BatchNorm + ReLU}$};

\node (conv2) [process, yshift=0.21cm,minimum width=10.88cm,minimum height=0.9cm,below of=batchnorm,inner sep=2.5pt]  {$\mathrm{Conv(1 \times 1, }F)$};
\node (batchnorm2) [process,minimum width=10.88cm,yshift=0.2cm,minimum height=1.03cm,,below of=conv2]  {$\mathrm{BatchNorm + ReLU}$};

\node at ($(batchnorm2)+(-0.5cm,-1.5cm)$) [font=\fontsize{26.5}{28.5}\selectfont] (labels) {$\mathbf{r}$};

\draw[arrow] 
  ++(conv1.north)+(0,1.5) 
  -- 
  (conv1.north) 
  node[midway, xshift=2.35cm, yshift=0.25cm,] 
  {${F \times \frac{H}{4} \times \frac{W}{4}}$};

\draw[arrow] (batchnorm2) -- ++(0,-2) node[midway, xshift=2.35cm, yshift=-0.25cm] {${F \times \frac{H}{8} \times \frac{W}{8}}$};

\end{tikzpicture}}
    {\footnotesize (a) Feature encoder ($\mathbf{FE}$)}
  \end{minipage}\hfill
  \begin{minipage}[t]{0.49\linewidth}
  \centering
  \hspace{-1cm}
    \resizebox{1.17\linewidth}{!}{\usetikzlibrary{shapes.geometric, arrows.meta, calc,backgrounds}

\definecolor{encoder_decoder}{RGB}{213, 232, 212}
\definecolor{rectangle}{RGB}{255, 245, 204}
\definecolor{bottleneck_color}{RGB}{237, 233, 112}
\definecolor{_decoder}{RGB}{255, 204, 255} 
\definecolor{conv}{RGB}{255, 255, 255}
\definecolor{quant_color}{RGB}{255, 162, 234}
 \definecolor{ae_color}{RGB}{221, 221, 221}
 \definecolor{_rectangle}{RGB}{221, 221, 221}
 \definecolor{source_color}{RGB}{200, 201, 164}
 \definecolor{hyper_color}{RGB}{255, 219, 77}

\tikzstyle{process} = [rectangle,minimum width=5cm, minimum height=1cm, text
centered, draw=black, text=black,line width = 1pt, fill=conv]
 \tikzstyle{arrow} = [-Triangle, line width=1pt]
 \tikzstyle{label} = [text width= 0.2cm, align=center]
 \tikzstyle{waypoint}=[fill,circle,minimum size=3.5pt,inner sep=0pt]
 \tikzstyle{encoder} = [trapezium,
    trapezium angle=55,
    trapezium stretches=true,
    minimum width=7.95cm,
    minimum height=5.2cm,
    trapezium right angle=90,
    trapezium left angle=90,
    shape border rotate=180,
    text centered,
    ]

  \tikzstyle{decoder} = [trapezium,
    trapezium angle=55,
    trapezium stretches=true,
    minimum width=8cm,
    minimum height=4.6cm,
    trapezium right angle=85,
    trapezium left angle=85,
    line width=1.5pt,
    shape border rotate=180,
    text centered,
    draw=black]

\begin{tikzpicture}[node distance = 1.5cm,font=\fontsize{24}{26}\selectfont]

\node (source_decoder) [draw, rectangle,minimum width=11.6cm, minimum height=5.65cm,   draw=none, fill=bottleneck_color]   {};

\node (sd_batchnorm) [process,xshift=0cm, yshift=0.42cm,minimum width=10.88cm,minimum height=1.03cm, above of=source_decoder]  {$\mathrm{BatchNorm + ReLU}$};

\node (sd_conv1) [process,xshift=0cm, yshift=0.18cm,minimum width=10.88cm,minimum height=0.9cm, below of=sd_batchnorm]  {$\mathrm{DWUpConv(3 \times 3,}F,G,\rho=2)$};

\node (sd_batchnorm2) [process,minimum width=10.88cm,yshift=0.2cm,minimum height=1.03cm,below of=sd_conv1]  {$\mathrm{BatchNorm + ReLU}$};

\node (sd_conv2) [process, yshift=0.22cm,minimum width=10.88cm,minimum height=0.9cm,below of=sd_batchnorm2,inner sep=2.5pt]  {$\mathrm{UpConv(1 \times 1, }F)$};

\draw[arrow] 
  ++(sd_conv2.south)+(0,-1.5) 
  -- 
  (sd_conv2.south) 
  node[midway, xshift=2.35cm, yshift=-0.25cm,] 
  {${F \times \frac{H}{8} \times \frac{W}{8}}$};

\draw[arrow] (sd_batchnorm) -- ++(0,2.1) node[midway, xshift=2.35cm, yshift=0.25cm]  {${F \times \frac{H}{4} \times \frac{W}{4}}$};

\node at ($(sd_conv2)+(-0.5cm,-1.5cm)$) [font=\fontsize{26.5}{28.5}\selectfont] {$\mathbf{\hat{r}}$};

\node at ($(sd_batchnorm)+(-0.5cm,1.45cm)$) [font=\fontsize{26.5}{28.5}\selectfont] {$\mathbf{\hat{z}}$};

\end{tikzpicture}}
    {\footnotesize (b) Feature decoder ($\mathbf{FD}$)}
  \end{minipage}
  \vspace{3pt}
  \caption{Network architecture of \textbf{feature encoder} $\mathbf{FE}$ and \textbf{feature decoder} $\mathbf{FD}$ utilized in Fig.\ \ref{fig:high_level}, taken from~\cite{ahuja2023neural}.}
  \label{fig:fe_fd}
\end{figure}
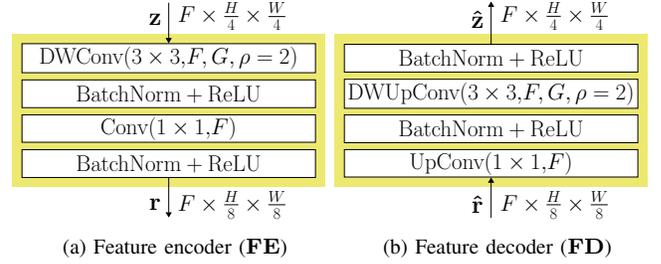

In Figure \ref{fig:ce_cd}, we show the detailed network architectures for the compression encoder $\mathbf{CE}$ and compression decoder $\mathbf{CD}$. Both $\mathbf{CE}$ and $\mathbf{CD}$ follow a hyperprior architecture \cite{balle2018variational,nazirjd}. The compression encoder $\bf{CE}$ with parameters $\bm{\theta}^{\mathrm{CE}}$ receives the latent representation $\mathbf{r}$ and produces a latent bitstream $\bf{b_{\hat{r}}=CE^{(r)}(r;\bm{\theta}^{\mathrm{CE}})}$ through a quantizer $\bf{Q}^{\mathcal{R}}$ and an arithmetic encoder $\bf{AE}^{\mathcal{R}}$ \cite{AE}. The elements of bitstream $\bf{b_{\hat{r}}}$ reveal a substantial amount of statistical dependencies and correlations. Accordingly, an enhancement bitstream $\bf{b_{\hat{h}}}$ is being introduced in $\mathbf{CE}$ to model a time-variant standard deviation vector $\bf{\boldsymbol{\sigma}}$ for arithmetic encoding $\bf{AE}^{\mathcal{H}}$\cite{balle2018variational}. It is produced by a quantizer $\bf{Q}^{\mathcal{H}}$ and a hyperprior encoder $\mathbf{HE}$ as $\bf{b_{\hat{h}}= CE^{(h)}(r;\bm{\theta}^{\mathrm{HE}})}$. Due to the introduction of $\bf{b_{\hat{h}}}$, each element in $\bf{\hat{r}}$ can now be modeled as zero-mean Gaussian with its individual standard deviation $\sigma_{j}$ using a hyperprior decoder $\bf{HD}$ as $\bf{\boldsymbol{\sigma}=(\sigma_{\textit{j}})=HD(\hat{h};\bm{\theta}^{\mathrm{HD}})}$ with parameters $\bm{\theta}^{\mathrm{HD}}$. The compression decoder $\mathbf{CD}$ with parameters $\bm{\theta}^{\mathrm{CD}}$ reconstructs the latent representation through arithmetic decoders $\bf{AD}^{\mathcal{R}}$ and $\bf{AD}^{\mathcal{H}}$ \cite{AE}, along with a hyperprior decoder $\bf{HD}$ as $\bf{\hat{r}}=\bf{CD}(\bf{b_{\hat{r}}},\bf{b_{\hat{h}}};\bm{\theta}^{\mathrm{CD}})$. Both, $\bf{CE}$ and $\bf{CD}$, contain convolutional and transposed convolutional layers, which are represented by $\mathrm{Conv(\textit{h} \times \textit{h},\textit{F}, \rho)}$ and $\mathrm{UpConv(\textit{h} \times \textit{h},\textit{F}, \rho)}$, respectively, where $\mathrm{\textit{h} \times \textit{h}}$ is the kernel size, $F$ is the number of kernels, and stride $\rho=2$.

The expected bitrate during training can be defined as

\begin{equation}
J^{\mathrm{rate}} = \mathbb{E}_{\bf{x} \sim \mathrm{p_{train}}}  \biggl[ \frac{  \mathrm{-log_{2}} (\mathrm{P_{\bf{\hat{r}}}}(\bf{\hat{r}} | \bf{\hat{h}}))  \mathrm{-log_{2}} (\mathrm{P_{\bf{\hat{h}}}}(\bf{\hat{h}}))  }{H \cdot W} \biggr],
\label{Eq:rate}
\end{equation}
with $\mathbb{E}_{\bf{x} \sim \mathrm{p_{train}} }$ representing the expectation over a minibatch in the dataset, $\mathrm{P_{\bf{\hat{r}}}}$ and $\mathrm{P_{\bf{\hat{h}}}}$ denoting the discrete probability distributions over the quantized latent spaces $\bf{\hat{r}}$ and $\bf{\hat{h}}$, respectively.

Since the task decoder $\bf{D}$ performs semantic segmentation, the distortion objective is formulated as a cross-entropy loss. 
\begin{equation}
J^{\mathrm{dist}} = \mathbb{E}_{\bf{x} \sim \mathrm{p_{train}}}  \left[ \frac{1}{|\mathcal{I}|}  \sum_{\substack{i \in \mathcal{I} }} \sum_{\substack{s \in \mathcal{S} }} \Bar{y}_{i,s}\cdot\mathrm{log}(y_{i,s}) \right].
\label{Eq:CE}
\end{equation}
Here, $ \Bar{\bm{y}}=(\Bar{y}_{i,s}) \in \{0,1 \}^{H \times W \times S}$ is the one-hot-encoded ground truth and we have $\forall i \in \mathcal{I}:\sum_{s \in \mathcal{S}}y_{i,s}=1$,\quad$\sum_{s \in \mathcal{S}}\Bar{y}_{i,s}=1$.~By combining (\ref{Eq:rate}) and (\ref{Eq:CE}), we obtain the RD trade-off in our total loss as follows: 
\begin{equation}
J = \alpha \cdot J^{\mathrm{dist}} + (1 - \alpha) \cdot  J^{\mathrm{rate}},
\label{Eq:RD}
\end{equation}
controlled by the hyperparameter $\alpha \in (0,1)$.

\subsection{PROPOSED JOINT FEATURE AND TASK DECODING FOR SEGDEFORMER}

Recent work \cite{nazirjd} in distributed semantic segmentation proposed joint source and task decoding for convolutional neural network (CNN)-based task decoders such as \texttt{DeepLabV3} \cite{deeplabv3}. Even though \texttt{DeepLabV3} is a strong distributed semantic segmentation baseline, it is not a SOTA method in general semantic segmentation, which is dominated by transformer-based task decoders such as  \texttt{SegDeformer} \cite{segdeformer}. However, \texttt{SegDeformer} incurs very high computational complexity due to its context mining block, making it unsuitable for both in-car and distributed applications. \textit{Therefore, to address these limitations, we propose joint feature and task decoding for \texttt{SegDeformer}, aiming to lower its computational complexity without compromising on the semantic segmentation performance in both in-car and distributed applications}.
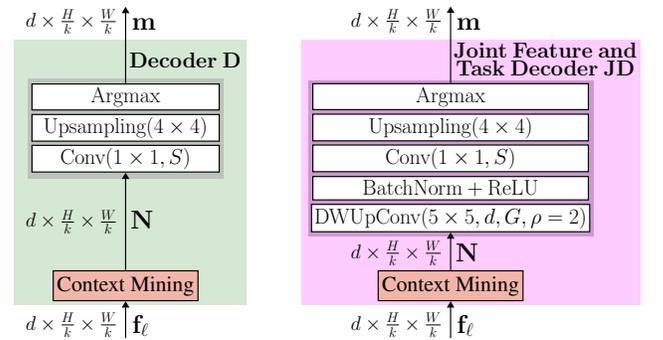
\begin{figure}[t!]
   \hspace{-3.35em}
  \resizebox{0.57\textwidth}{!}{\definecolor{encoder_decoder}{RGB}{213, 232, 212}
\definecolor{rectangle}{RGB}{255, 245, 204}
\definecolor{bottleneck_color}{RGB}{230, 221, 184}
\definecolor{_decoder}{RGB}{255, 204, 255} 
\definecolor{conv}{RGB}{255, 255, 255}
\definecolor{quant_color}{RGB}{232, 123, 142}
\definecolor{ae_color}{RGB}{221, 221, 221}
\definecolor{_rectangle}{RGB}{221, 221, 221}
\definecolor{source_color}{RGB}{200, 201, 164}
\definecolor{upsampling_block}{RGB}{204, 154, 204}
\definecolor{ce_color}{RGB}{235, 245, 235}
\definecolor{simple_attention}{RGB}{177, 209, 227}
\definecolor{cross_attention}{RGB}{242, 182, 169}
\definecolor{seg_block}{RGB}{188, 196, 188}
\definecolor{jd}{RGB}{255, 204, 255} 
\definecolor{context_mining}{RGB}{242, 182, 169}
\tikzstyle{process} = [rectangle,minimum width=5cm, text
centered, draw=black, text=black,line width = 1pt, fill=conv]
\tikzstyle{arrow} = [-Triangle, line width=1pt]
\tikzstyle{label} = [text width= 0.2cm, align=center]
\tikzstyle{waypoint}=[fill,circle,minimum size=3.5pt,inner sep=0pt]

\begin{tikzpicture} [node distance = 1.5cm,font=\fontsize{18.5}{20.5}\selectfont]

\coordinate (left_origin) at (-6.2,0);
\coordinate (right_origin) at (4.05,0);

\node (d) [draw, rectangle, minimum width=7cm, minimum height=8.cm, line width=0pt, draw=none, fill=encoder_decoder, anchor=center] at (left_origin) {};

\node (feature_tokens) [below=1cm of d.south,xshift = -0.15cm,yshift=0.cm]  {};
\node (feature_tokens_label) [right of=feature_tokens,xshift = -1.05cm,yshift=0.5cm,font=\fontsize{22}{24}\selectfont]  {$\mathbf{f}_{\ell}$};

\node (context_mining) [process, fill = context_mining,above of=feature_tokens, xshift=0.cm, yshift=0.2cm,  minimum width=3cm, minimum height=0.9cm ] {\shortstack{$\text{Context Mining}$}};

\node (N) [above of=context_mining,xshift = 0.5cm,yshift=0.5cm,font=\fontsize{22.25}{24.25}\selectfont]  {$\mathbf{N}$};

\node (output_block) [ above of =context_mining, minimum height=3.cm, minimum width=5.88cm,  xshift=0cm, yshift=3.25cm,  fill=seg_block] { } ;

\node (decoder_label) [minimum width=1cm,xshift=-1.15cm,yshift=-1.4cm, minimum height=0.9cm,above= of output_block.north east]  {\shortstack{$\mathbf{Decoder}$  $\mathbf{D}$}};

\node (argmax) [process, minimum width=5.64cm,minimum height=0.7cm, xshift=0cm,yshift=0.94cm,inner sep=3pt ] at (output_block)   {$\mathrm{Argmax}$};
\node (upsampling) [process, below of=argmax, minimum width=5.64cm,minimum height=0.7cm, xshift=0cm,yshift=0.57cm,inner sep=2.2pt ]    {$\mathrm{Upsampling (4 \times 4)}$};
\node (conv_seg) [process, below of=upsampling, minimum width=5.64cm,minimum height=0.7cm,inner sep=2.2pt, xshift=0cm,yshift=0.57cm]    {$\mathrm{Conv(1\times1},S)$};

\draw[arrow] (argmax) -- ++(0,2.72) node[midway, xshift=0.55cm, yshift=0.63cm,font=\fontsize{22}{24}\selectfont] {$\mathbf{m}$};


\node at ($(argmax)+(-0.5cm,0.43cm)$) [ xshift=-1.13cm,yshift=1.85cm,font = \LARGE] (labels) {${d \times \frac{H}{k} \times \frac{W}{k} }$};

\draw [arrow] (feature_tokens) to  node[midway, above, align=center,xshift=-1.63cm,yshift=-0.65cm,font = \LARGE ] {${d \times \frac{H}{k} \times \frac{W}{k} }$}    (context_mining);
\draw[arrow] (context_mining)  to  node[midway, above, align=center,xshift=-1.63cm,yshift=-0.51cm,font = \LARGE ] {${d \times \frac{H}{k} \times \frac{W}{k} }$}     (conv_seg);

\node[align=center, anchor=north,yshift=-0.48cm, font=\fontsize{17.85}{19.85}\selectfont] 
  at ($(d.south) + (0,-1)$)
  {\parbox{8cm}{
    (a) Network architecture of \textbf{task decoder} $\mathbf{D}$ utilized in Fig.\ 1(a), adapted from \cite{segdeformer}.
  }};

 \node (jd) [draw, rectangle, minimum width=10.2cm, minimum height=8cm, line width=0pt, draw=none, fill=jd, anchor=center] at (right_origin) {};
 
\node (feature_tokens) [below=1cm of jd.south,xshift = -0.61cm,yshift=0.cm]  {};
\node (feature_tokens_label) [right of=feature_tokens,xshift = -1.05cm,yshift=0.5cm,font=\fontsize{22}{24}\selectfont]  {$\mathbf{f}_{\ell}$};

\node (context_mining) [process, fill = context_mining,above of=feature_tokens, xshift=0.cm, yshift=0.2cm,  minimum width=3cm, minimum height=0.9cm ] {\shortstack{$\text{Context Mining}$}};

\node (N) [above of=context_mining,xshift = 0.5cm,yshift=-0.5cm,font=\fontsize{22.25}{24.25}\selectfont]  {$\mathbf{N}$};

\node (output_block) [ above of =context_mining, minimum height=4.7cm, minimum width=8.6cm,  xshift=0cm, yshift=2.35cm,  fill=upsampling_block] { } ;

\node (argmax) [process, minimum width=8.3cm,minimum height=0.7cm, xshift=0cm,yshift=1.85cm,inner sep=3pt ] at (output_block)   {$\mathrm{Argmax}$};
\node (upsampling) [process, below of=argmax, minimum width=8.3cm,minimum height=0.7cm, xshift=0cm,yshift=0.57cm,inner sep=2.2pt ]    {$\mathrm{Upsampling (4 \times 4)}$};
\node (conv_seg) [process, below of=upsampling, minimum width=8.3cm,minimum height=0.7cm,inner sep=2.2pt, xshift=0cm,yshift=0.57cm ]    {$\mathrm{Conv(1\times1},S)$};
 
\node (bn_relu_upconv) [process, minimum width=8.3cm,minimum height=0.7cm, xshift=0.0cm,yshift=0.6cm,below of=conv_seg,inner sep=3pt]   {$\mathrm{BatchNorm + ReLU}$};
\node (upconv) [process, minimum width=8.3cm,minimum height=0.4cm, xshift=0.0cm,yshift=0.6cm,below of=bn_relu_upconv,inner sep=2.5pt]   {$\mathrm{DWUpConv (5 \times 5}, d, G, \rho=2)$};

\node (decoder_label) [ minimum width=3cm,  minimum height=0.9cm,  xshift=-1.45cm, yshift=-1.5cm, above=of output_block.north east] 
{\shortstack{$\mathbf{Joint \ Feature \ and}$ \\ $\mathbf{Task \  Decoder \ JD}$}};
\draw[arrow] (argmax) -- ++(0,2.72) node[midway, xshift=0.55cm, yshift=0.63cm,font=\fontsize{22}{24}\selectfont] {$\mathbf{m}$};
\node at ($(argmax)+(-0.5cm,0.43cm)$) [ xshift=-1.13cm,yshift=1.85cm,font = \LARGE] (labels) {${d \times \frac{H}{k} \times \frac{W}{k} }$};

\draw [arrow] (feature_tokens) to  node[midway, above, align=center,xshift=-1.63cm,yshift=-0.65cm,font = \LARGE ] {${d \times \frac{H}{k} \times \frac{W}{k} }$}    (context_mining);

\draw[arrow] (context_mining)  to  node[midway, above, align=center,xshift=-1.63cm,yshift=-0.55cm,font = \LARGE] {${d \times \frac{H}{k} \times \frac{W}{k} }$}     (upconv);   

\node (caption)[anchor=north, font=\fontsize{17.85}{19.85}\selectfont] at ([xshift=0.3cm,yshift=-1.5cm]jd.south) 
{\parbox[c]{11.cm}{\raggedright 
(b) \textbf{Proposed} network architecture of the \textbf{joint feature and task decoder} $\mathbf{JD}$,
uti-\\lized in Fig.\ 1(b).}};

\node (space) [below of=caption,yshift=0cm]{};

\end{tikzpicture}}
    \caption{Detailed overview of the decoder architectures for in-car and distributed semantic segmentation applications. Fig.\ \ref{fig:overview_architectures}(a) shows \textbf{baseline decoder} $\mathbf{D}$, taken from \cite{segdeformer}. Fig.\ \ref{fig:overview_architectures}(b) illustrates the \textbf{proposed architecture} for the \textbf{joint feature and task decoder} $\mathbf{JD}$.}
    \label{fig:overview_architectures}
\end{figure}
Fig.\ \ref{fig:high_level}\subref{fig:ours_high_level} presents the \textit{proposed transformer-based} \textsf{in-car JD}, where the joint feature and task decoding utilizes latent representation $\bf{r}$ $\in \mathbb{R}^{F \times \frac{H}{8} \times \frac{W}{8}}$ instead of bottleneck features $\bf{z}$ $\in \mathbb{R}^{F \times \frac{H}{4} \times \frac{W}{4}}$ as an input to the proposed in-car joint feature and task decoder $\bf{JD}$ with parameters $\bm{\theta}^{\mathrm{JD}}$ to generate semantic segmentation map $\bf{m}=\bf{JD(r;\bm{\theta}^{\mathrm{JD}})}$, resulting in lower computational complexity than $\mathbf{D}$. Fig.\ \ref{fig:high_level}\subref{fig:ours_high_level} also depicts the \textit{proposed transformer-based} \textsf{distributed JD}, which omits the reconstruction of $\bf{\hat{z}}$ and redirects the quantized latent representation $\bf{\hat{r}}$ to the proposed distributed joint feature and task decoder $\bf{JD}$ with parameters $\bm{\theta}^{\mathrm{JD}}$ to produce a client-sided semantic segmentation mask as $\bf{m}=\bf{JD(\hat{r};\bm{\theta}^{\mathrm{JD}})}$\cite{nazirjd}. Similar to $\bf{D}$, the network architectures of both in-car and distributed task decoders $\bf{JD}$ in Fig.\ \ref{fig:high_level}\subref{fig:ours_high_level} are exactly the same. However, their weights $\bm{\theta}^{\mathrm{JD}}$ differ from each other.


In Figure~\ref{fig:overview_architectures}, the detailed architectures for the task decoder $\mathbf{D}$ and the proposed transformer-based joint feature and task decoder $\mathbf{JD}$ are illustrated. Fig.\ \ref{fig:overview_architectures}\subref{fig:baseline} shows the network architecture of $\mathbf{D}$ for both \textsf{in-car baseline} and \textsf{distributed baseline}. Similarly, Fig.\ \ref{fig:overview_architectures}(b) illustrates the network architecture of the proposed transformer-based joint feature and task decoder for both \textsf{in-car JD} and \textsf{distributed JD}.

The task decoder $\mathbf{D}$ receives the input features $\mathbf{f}_{\ell} \in \mathbb{R}^{d \times \frac{H}{k} \times \frac{W}{k}}$ in a pre-defined layer $\ell$ of the network, with internal dimension $d=256$, and downsampling factor $k=4$. They correspond to the bottleneck feature tokens $\bf{f}_{\ell} = \bf{z}$ for the in-car application, and the reconstructed feature tokens $\bf{f}_{\ell} = \bf{\hat{z}}$ for the distributed application. Similarly, the proposed joint decoder $\mathbf{JD}$ also receives the input features $\mathbf{f}_{\ell} \in \mathbb{R}^{d \times \frac{H}{k} \times \frac{W}{k}}$ in layer $\ell$ of the network with internal dimension $d=48$, and downsampling factor $k=8$. They correspond to the bottleneck feature tokens $\bf{f}_{\ell} = \bf{r}$ for the in-car application, and the reconstructed feature tokens $\bf{f}_{\ell} = \bf{\hat{r}}$ for the distributed application. Note that, except for $\mathbf{f}_{\ell}$, everything else is exactly the same for both applications.

\begin{figure}[t!]
  \hspace{-4em}
  \resizebox{1.16\linewidth}{!}{\usetikzlibrary{positioning,shapes,arrows,arrows.meta,fit,backgrounds,calc}

\definecolor{encoder_decoder}{RGB}{242, 182, 169}
\definecolor{rectangle}{RGB}{255, 245, 204}
\definecolor{bottleneck_color}{RGB}{230, 221, 184}
\definecolor{_decoder}{RGB}{255, 204, 255} 
\definecolor{conv}{RGB}{255, 255, 255}
\definecolor{quant_color}{RGB}{232, 123, 142}
\definecolor{ae_color}{RGB}{221, 221, 221}
\definecolor{_rectangle}{RGB}{221, 221, 221}
\definecolor{source_color}{RGB}{200, 201, 164}
\definecolor{hyper_color}{RGB}{255, 219, 77}
\definecolor{ce_color}{RGB}{235, 245, 235}
\definecolor{simple_attention}{RGB}{177, 209, 227}
\definecolor{cross_attention}{RGB}{255,163,87}
\definecolor{seg_block}{RGB}{188, 196, 188}

\tikzstyle{process} = [rectangle,minimum width=5cm, text
centered, draw=black, text=black,line width = 1pt, fill=conv]
\tikzstyle{arrow} = [-Triangle, line width=1pt]
\tikzstyle{label} = [text width= 0.2cm, align=center]
\tikzstyle{waypoint}=[fill,circle,minimum size=3.5pt,inner sep=0pt]

\newcommand{\Plus}{%
    \tikz[baseline=-0.5ex, line width=1, scale=0.2]{
        \draw[black, fill=black] (0,-1) -- (0,1);  
        \draw[black, fill=black] (-1,0) -- (1,0);  
    }%
}
\begin{tikzpicture}[node distance = 1.5cm,font=\fontsize{18.25}{20.25}\selectfont]

\node (background) [draw, rectangle, minimum width=15.75cm, minimum height=5.75cm, line width=0pt, draw=none, fill=encoder_decoder, anchor=center] at (0,0) {};

\node (feature_tokens) [below=0.8cm of background.south,xshift = -0.7cm,yshift=0.cm]  {};
\node (feature_tokens_label) [right of=feature_tokens,xshift = -1.13cm,yshift=0.5cm]  {$\mathbf{f}_{\ell}$};
\node (self_attention) [process, fill = simple_attention,above of=feature_tokens, xshift=0.cm, yshift=1cm,  minimum width=3cm, minimum height=0.9cm,font = \LARGE] {\shortstack{$\text{Single-head}$ \\ $\text{Self Attention}$}};
\node (skip_conv) [process, minimum width=3.5cm,minimum height=0.1cm, xshift=-2.95cm,yshift=-0.01cm,left of =self_attention,inner sep=3pt,font = \LARGE]  {$\mathrm{Conv(1\times1},F)$};
\node (custom_cross_attention) [process, fill = cross_attention,right of=self_attention, xshift=3.6cm, yshift=0cm,  minimum width=3cm, minimum height=0.9cm,font = \LARGE] {\shortstack{$\text{Single-head Custom}$ \\ $\text{Cross Attention}$}};

\node (cross_attention) [process, fill = simple_attention,above of=custom_cross_attention, xshift=0.0cm, yshift=0.6cm,  minimum width=2cm, minimum height=0.9cm,font = \LARGE] {\shortstack{$\text{Single-head Self Attention}$}};

\node (add) [above of=cross_attention, draw=black, line width=1pt, fill=conv, xshift=-5.1cm, yshift=0.1cm, shape=circle, minimum size=0.8cm, inner sep=0pt] {$\Plus$};

\node (context_mining_label) [minimum width=1cm,xshift=2cm,yshift=-1.3cm, minimum height=0.9cm,above= of cross_attention]  {\shortstack{$\mathbf{Context}$ \\ $\mathbf{Mining}$}};

\node (way)[waypoint] at ($(feature_tokens)!0.43!(self_attention)$) {};
\node (way2) [waypoint, left of=way, xshift=-0.15cm, yshift=0cm] {};
\node (way3) [waypoint, right of=way, xshift=0.17cm,] {};
\node (way4) [waypoint, xshift=0cm] at ($(custom_cross_attention)!0.43!(cross_attention)$) {}; 
\node (way5) [waypoint, right of=way3, xshift=-0.95cm,] {};
\draw [line width=1,font = \Large] (feature_tokens) to  node[midway, above, align=center,xshift=-1.39cm,yshift=-0.48cm,font = \Large] {${d \times \frac{H}{k} \times \frac{W}{k} }$}    (way);

\draw[arrow] (way3) to  node[midway, above, align=center,xshift=-0.5cm,yshift=-0.cm] {$\mathbf{V}_{\ell}$}  ++(0,0)  --  ([xshift=-0.185cm]self_attention.south east);
\draw [arrow]  (way) to  node[midway, above, align=center,xshift=-0.52cm,yshift=-0.42cm] {$\mathbf{K}_{\ell}$}  (self_attention);
\draw[arrow] (way2) to  node[midway, above, align=center,xshift=-0.5cm,yshift=-0.45cm] {$\mathbf{Q}_{\ell}$}   ([xshift=0.2cm]self_attention.south west);

\draw [line width=1] (way2) to  node[midway, above] {}    (way);
\draw [line width=1] (way) to  node[midway, above] {}    (way3);

\draw [arrow]  (way3) to node[pos=0.5, above, xshift=-0.65cm] {}  ($(way3) + (3.45,0.)$) -- ( custom_cross_attention);
\draw [arrow] (way2) to  node[midway, xshift=-2.2cm, yshift=1.2cm] {} ++(-2.82,0) --  (skip_conv);
\draw [line width=1] (custom_cross_attention) --  (way4);

\draw [arrow] (way4) to node[pos=0.5,yshift=-0.01cm,xshift=0.79cm, above] {$\mathbf{V}^{(\mathbf{H})}$} ($(way4) + (3.18,0)$) -- ([xshift=-0.2cm] cross_attention.south east);
\draw [arrow] (way4) to node[pos=0,xshift=-0.77cm,yshift=-0.08cm, above] {$\mathbf{K}^{(\mathbf{H})}$} ([xshift=0cm]cross_attention.south);

\draw [arrow] (way5) to  node[midway, xshift=-0.5cm, yshift=1.38cm] {$\mathbf{Q}_{\ell}$} ++(0.,2.5) -- ++(0.,0) --  ([xshift=0.495cm] cross_attention.south west);

\draw [arrow] (cross_attention) to  node[midway, xshift=-1.37cm, yshift=0cm,font = \Large] {${d \times \frac{H}{k} \times \frac{W}{k} }$} ++(0,1.6) --  (add.east);
\draw [arrow] (self_attention) to  node[midway, xshift=-1.37cm, yshift=0cm,font = \Large]  {${d \times \frac{H}{k} \times \frac{W}{k} }$}   (add);
\draw [arrow] (skip_conv) to  node[midway, xshift=-1.37cm, yshift=-0.09cm, font = \Large]  {${d \times \frac{H}{k} \times \frac{W}{k} }$} ++(0,3.7) --++(2.5,0) --  (add.west);

\draw[arrow] (add) -- ++(0,1.25) node[midway, xshift=0.45cm, yshift=0.03cm,] {$\mathbf{N}$};
\node at ($(add)+(-0.5cm,0.8cm)$) [ xshift=-0.9cm,yshift=0.06cm,font = \Large] (labels) {${d \times \frac{H}{k} \times \frac{W}{k} }$};
\end{tikzpicture}
}
    \caption{\textbf{Context mining} block utilized in Fig.\ \ref{fig:overview_architectures}, adapted from \cite{segdeformer}.  }
    \label{fig:context_mining}
\end{figure}
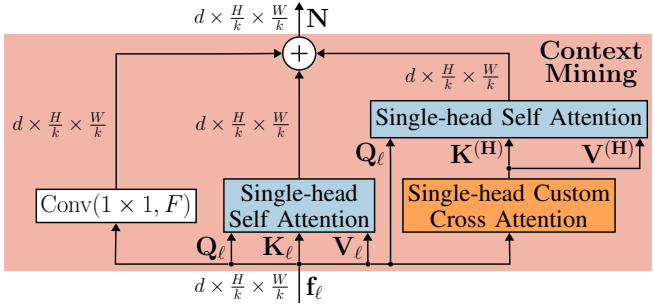

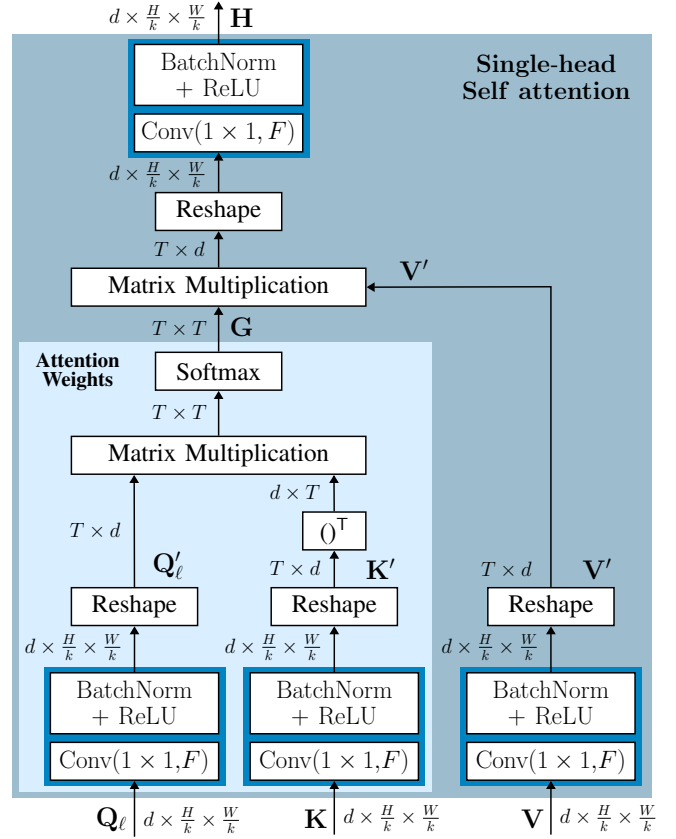
\begin{figure}[t!]
  \hspace{-3.2em}
  \resizebox{1.15\linewidth}{!}{ \usetikzlibrary{positioning,shapes,arrows,arrows.meta,fit,backgrounds,calc}
\definecolor{rectangle}{RGB}{255, 245, 204}
\definecolor{conv}{RGB}{255, 255, 255}
\definecolor{ce_color}{RGB}{235, 245, 235}
\definecolor{simple_attention}{RGB}{177, 209, 227}
\definecolor{self_attention}{RGB}{157,189,207}
\definecolor{seg_block}{RGB}{188, 196, 188}
\definecolor{attention_block}{RGB}{217,239,255}

\definecolor{query_color}{RGB}{0, 132, 194}
\definecolor{key_color}{RGB}{199,199, 199}
\definecolor{value_color}{RGB}{0, 132, 194} 

\tikzstyle{process} = [rectangle,minimum width=5cm, minimum height=1cm, text
centered, draw=black, text=black,line width = 1pt, fill=conv]
 \tikzstyle{arrow} = [-Triangle, line width=1pt]
 \tikzstyle{label} = [text width= 0.2cm, align=center]
 \tikzstyle{waypoint}=[fill,circle,minimum size=3.5pt,inner sep=0pt]

\newcommand{\CircularCross}{
 	\tikz[baseline=0.25ex, line width=1, scale=0.25]{
        	\draw[thick] (-0.5,-0.5) -- (0.5,0.5); 
        	\draw[thick] (-0.5,0.5) -- (0.5,-0.5); 
	}
}

\begin{tikzpicture}[node distance = 1.5cm,font=\LARGE]

\node (point) at (0,0) {};
\node (background) [ rectangle, minimum width=15.2cm, minimum height=18.17cm,  line width=0pt,draw=none, fill=self_attention] at (point)  {};
\node (attention_block) [above=0.12cm of background.south,rectangle,  xshift=-2.52cm, minimum width=9.8cm, minimum height=10.7cm, line width=1pt,draw=none, fill=attention_block]    {};

\node (proj_query) [above=0.01cm of attention_block.south west, minimum width=4.31cm, xshift=2.75cm, yshift=0.15cm, minimum height=2.81cm,draw=none, fill=query_color]    {};
 \node (conv_query) [process, xshift=0cm, yshift=-0.83cm,  minimum width=4.cm, minimum height=0.9cm] at (proj_query.center) {$\mathrm{Conv(1\times 1,}F)$};
\node (bn_relu_query) [process, xshift=0cm, yshift=-0.12cm,  minimum width=4cm, minimum height=1.45cm,above of=conv_query] {\shortstack{$\mathrm{BatchNorm}$ \\[0.17em]$\mathrm{ + \ ReLU}$}};
\node (reshape_query) [process, xshift=0cm, yshift=0.8cm,  minimum width=3cm, minimum height=0.9cm,above of=bn_relu_query]   {\shortstack{$\text{Reshape}$}};

\node (q_prime) [ xshift=0.8cm, yshift=-0.52cm,  above of=reshape_query,font=\fontsize{18.25}{20.25}\selectfont]   {\shortstack{$\mathbf{Q}_{\ell}^{\prime}$}};
 
 \node (proj_key) [right of=proj_query, minimum width=4.31cm, xshift=3.25cm, yshift=0cm, minimum height=2.81cm,  line width=0pt,draw=none, fill=query_color]    {};
\node (conv_key) [process, xshift=0cm, yshift=-0.83cm,  minimum width=4cm, minimum height=0.9cm] at (proj_key.center) {$\mathrm{Conv(1\times 1,}F)$};
\node (bn_relu_key) [process, xshift=0cm, yshift=-0.12cm,  minimum width=4cm, minimum height=1.45cm,above of=conv_key] {\shortstack{$\mathrm{BatchNorm}$ \\[0.17em] $\mathrm{ + \ ReLU}$}};

 \node (proj_value) [right of=proj_key, minimum width=4.31cm, xshift=3.65cm, yshift=0cm, minimum height=2.81cm,draw=none, fill=value_color]    {};
 \node (conv_value) [process, xshift=0cm, yshift=-0.83cm,  minimum width=4.cm, minimum height=0.9cm] at (proj_value.center) {$\mathrm{Conv(1\times 1,}F)$};
\node (bn_relu_value) [process, xshift=0cm, yshift=-0.12cm,  minimum width=4cm, minimum height=1.45cm,above of=conv_value]  {\shortstack{$\mathrm{BatchNorm}$ \\[0.17em] $\mathrm{ + \ ReLU}$}};
\node (reshape_value) [process, xshift=0cm, yshift=0.8cm,  minimum width=3cm, minimum height=0.9cm,above of=bn_relu_value]   {\shortstack{$\text{Reshape}$}};
\node (v_prime) [ xshift=1.1cm, yshift=-0.6cm,  above of=reshape_value,font=\fontsize{18.25}{20.25}\selectfont]   {\shortstack{$\mathbf{V}^{\prime}$}};
\node (query) [below of = proj_query,xshift = 0cm,yshift=-1.25cm]  {};
\node (key) [ below of = proj_key,xshift = 0cm,yshift=-1.2cm]  {};
\node (value) [ below of = proj_value,xshift = 0cm,yshift=-1.2cm]  {};
\node (reshape_key) [process, xshift=0cm, yshift=0.8cm,  minimum width=3cm, minimum height=0.9cm,above of=bn_relu_key]   {\shortstack{$\text{Reshape}$}};
\node (k_prime) [ xshift=1.1cm, yshift=-0.6cm,  above of=reshape_key,font=\fontsize{18.25}{20.25}\selectfont]   {\shortstack{$\mathbf{K}^{\prime}$}};
\node (transpose) [process, above of = reshape_key,xshift=0.cm, yshift=0.3cm,  minimum width=1.5cm, minimum height=0.9cm]  {\shortstack{$\text{()}^{\mathsf{T}}$}};
 \node (matrix_mul) [process, xshift=-2.75	cm, yshift=0.3cm,  minimum width=7cm, minimum height=0.9cm,above of=transpose]   {\shortstack{$\text{Matrix Multiplication}$}};
 
\node (softmax) [process, xshift=0cm, yshift=0.5cm,  minimum width=3cm, minimum height=0.9cm,above of=matrix_mul]   {\shortstack{$\text{Softmax}$}};
\node (g) [ xshift=0.55cm, yshift=-0.45cm,  above of=softmax,font=\fontsize{18.25}{20.25}\selectfont]   {\shortstack{$\mathbf{G}$}};
\node (matrix_mul2) [process, xshift=-0cm, yshift=0.5cm,  minimum width=7cm, minimum height=0.9cm,above of=softmax]   {\shortstack{$\text{Matrix Multiplication}$}};
 \node (v_prime) [ xshift=3.15cm, yshift=0.41cm,  right of=matrix_mul2,font=\fontsize{18.25}{20.25}\selectfont]   {\shortstack{$\mathbf{V}^{\prime}$}};

\node (reshape_matmul) [process, xshift=0cm, yshift=0.3cm,  minimum width=3cm, minimum height=0.9cm,above of=matrix_mul2]   {\shortstack{$\text{Reshape}$}};

\node (proj_ffn) [above of=reshape_matmul,rectangle, minimum width=4.31cm, xshift=0cm, yshift=1.2cm, minimum height=2.81cm,  draw=none, fill=value_color]    {};
\node (single_head_attention_label) [minimum width=3cm, minimum height=0.9cm, xshift=2cm, yshift=-0.9cm, right=of proj_ffn.north east ] {\shortstack{$\mathbf{Single}$$\text{-}$$\mathbf{head}$ \\ $\mathbf{Self}$$\ \mathbf{attention}$}};
\node (conv_ffn) [process, xshift=0cm, yshift=-0.83cm,  minimum width=4.cm, minimum height=0.9cm] at (proj_ffn.center) {$\mathrm{Conv(1\times1},F)$};
\node (bn_relu_ffn) [process, xshift=0cm, yshift=-0.12cm,  minimum width=4cm, minimum height=1.45cm,above of=conv_ffn]  {\shortstack{$\mathrm{BatchNorm}$\\[0.17em] $\mathrm{ + \ ReLU}$}};
 \node (h) [ xshift=0.55cm, yshift=0.41cm,  above of=proj_ffn,font=\fontsize{18.25}{20.25}\selectfont]   {\shortstack{$\mathbf{H}$}};
\node (label) [minimum width=3cm,xshift=1.37cm,yshift=0.52cm, minimum height=0.9cm, font = \Large] at (query)  {${d \times \frac{H}{k} \times \frac{W}{k}}$};
\node (label2) [minimum width=3cm,xshift=1.37cm,yshift=0.52cm, minimum height=0.9cm, font = \Large] at (key)  {${d \times \frac{H}{k} \times \frac{W}{k}}$};
 \node (label3) [minimum width=3cm,xshift=1.37cm,yshift=0.52cm, minimum height=0.9cm, font = \Large] at (value)  {${d \times \frac{H}{k} \times \frac{W}{k}}$};
\node (label4) [minimum width=3cm,xshift=-1.85cm,yshift=0.0cm, minimum height=0.9cm,font=\Large,left of=softmax]  {\shortstack{$\textbf{Attention}$ \\ $\textbf{Weights}$}};

\draw [arrow] (query) to node[midway, xshift=-0.55cm, yshift=-0.25cm,font=\fontsize{18.25}{20.25}\selectfont] {$\mathbf{Q}_{\ell}$} (conv_query);

\draw [arrow] (key) to  node[midway, xshift=-0.43cm, yshift=-0.25cm,font=\fontsize{18.25}{20.25}\selectfont]  {$\mathbf{K}$} (conv_key);
\draw [arrow] (value) to  node[midway, xshift=-0.43cm, yshift=-0.25cm,font=\fontsize{18.25}{20.25}\selectfont]  {$\mathbf{V}$} (conv_value);

 \draw [arrow] (bn_relu_query) to  node[midway, xshift=-1.45cm, yshift=0.05cm,font = \Large]  {${d \times \frac{H}{k} \times \frac{W}{k}}$} (reshape_query);
\draw [arrow] (bn_relu_key) to  node[midway, xshift=-1.45cm, yshift=0.05cm,font = \Large]  {${d \times \frac{H}{k} \times \frac{W}{k}}$} (reshape_key);
\draw [arrow] (bn_relu_value) to  node[midway, xshift=-1.45cm, yshift=0.05cm,font = \Large]  {${d \times \frac{H}{k} \times \frac{W}{k}}$} (reshape_value);
\draw [arrow] (reshape_key) to  node[midway, xshift=-0.93cm, yshift=0cm,font = \Large]  {${T  \times d}$} (transpose);
\draw [arrow] (transpose) to  node[midway, xshift=-0.93cm, yshift=0cm,font = \Large]  {${d  \times T}$} ([xshift=-0.775cm] matrix_mul.south east);
\draw [arrow] (reshape_query) to  node[midway, xshift=-0.93cm, yshift=0cm,font = \Large]  {${T  \times d}$} ([xshift=1.51cm] matrix_mul.south west);
\draw [arrow] (matrix_mul) to  node[midway, xshift=-0.93cm, yshift=0cm,font = \Large]  {${T  \times T}$} (softmax);
\draw [arrow] (softmax) to  node[midway, xshift=-0.93cm, yshift=0cm,font = \Large]  {${T  \times T}$} (matrix_mul2);
\draw[arrow] (reshape_value) to  node[midway, above, align=center,xshift=-1.06cm,yshift=-3.48cm,font = \Large] {${T  \times d}$}  ++(0,7.61)  --  (matrix_mul2.east);
\draw [arrow] (matrix_mul2) to  node[midway, xshift=-0.93cm, yshift=0cm,font = \Large]  {${T  \times d}$} (reshape_matmul);
\draw [arrow] (reshape_matmul) to  node[midway, xshift=-1.45cm, yshift=-0.05cm,font = \Large]  {${d \times \frac{H}{k} \times \frac{W}{k}}$} (conv_ffn);

\draw[arrow] (bn_relu_ffn) -- ++(0,1.75) node[midway, xshift=-1.45cm, yshift=0.12cm,font = \Large] {${d \times \frac{H}{k} \times \frac{W}{k} }$};

 \end{tikzpicture}}
    \caption{\textbf{Single-head self attention} block used in Fig.\ \ref{fig:context_mining}.}
    \label{fig:naive}
\end{figure}

\begin{figure}[t!]
   \hspace{-0.5em}
  \resizebox{0.45\textwidth}{!}{ \usetikzlibrary{positioning,shapes,arrows,arrows.meta,fit,backgrounds,calc}
 
\definecolor{encoder_decoder}{RGB}{213, 232, 212}
\definecolor{rectangle}{RGB}{255, 245, 204}
\definecolor{bottleneck_color}{RGB}{230, 221, 184}
\definecolor{_decoder}{RGB}{255, 204, 255} 
\definecolor{conv}{RGB}{255, 255, 255}
\definecolor{quant_color}{RGB}{232, 123, 142}
\definecolor{ae_color}{RGB}{221, 221, 221}
\definecolor{_rectangle}{RGB}{221, 221, 221}
\definecolor{source_color}{RGB}{200, 201, 164}
\definecolor{hyper_color}{RGB}{255, 219, 77}
\definecolor{ce_color}{RGB}{235, 245, 235}
\definecolor{simple_attention}{RGB}{177, 209, 227}
\definecolor{cross_attention}{RGB}{255,163,87}
\definecolor{seg_block}{RGB}{188, 196, 188}
\definecolor{attention_block}{RGB}{253,211,177}

\tikzstyle{process} = [rectangle,minimum width=5cm, minimum height=1cm, text
centered, draw=black, text=black,line width = 1pt, fill=conv]
 \tikzstyle{arrow} = [-Triangle, line width=1pt]
 \tikzstyle{label} = [text width= 0.2cm, align=center]
 \tikzstyle{waypoint}=[fill,circle,minimum size=3.5pt,inner sep=0pt]
\tikzstyle{data} = [
    draw, 
    cylinder, 
    line width=1pt, 
    shape border rotate=90,
    cylinder uses custom fill, 
    aspect=0.25,
    cylinder body fill=white,
    cylinder end fill=white,
    minimum height=15pt,
    minimum width=30pt,
    outer sep=0pt, 
    inner sep=4pt,
    align=center
]

\newcommand{\CircularCross}{
 	\tikz[baseline=0.25ex, line width=1, scale=0.25]{
        	\draw[thick] (-0.5,-0.5) -- (0.5,0.5); 
        	\draw[thick] (-0.5,0.5) -- (0.5,-0.5); 
	}
}

\begin{tikzpicture}[node distance = 1.5cm,font=\LARGE]

\node (fake) [ rectangle, minimum width=10.8cm, minimum height=4.8cm,  line width=0pt,draw=none, fill=cross_attention,yshift=0cm]   {};
\node (background) [ rectangle, below of=fake,minimum width=10.8cm, minimum height=17.cm,  line width=0pt,draw=none, fill=cross_attention,yshift=0cm]   {};

\node (feature_tokens) [below=0.75cm of background.south west,xshift = 3.5cm,yshift=0cm]  {};
\node (feature_tokens_lable) [above of=feature_tokens,xshift = 0.5cm,yshift=-1.05cm]  {$\mathbf{f}_{\ell}$};

\node (class_tokens) [data, above=0.0cm of background.south east,xshift = -2.5cm,yshift=0.4cm]  {$\mathbf{c=K}^{(\mathbf{c})}$};

\node (reshape) [process,minimum width=3cm,yshift=0.001cm, minimum height=0.9cm,above of=feature_tokens]  {$\text{Reshape}$};
\node (concat) [process,minimum width=3cm,xshift=0.cm,yshift=0.5cm, minimum height=0.9cm,above of=reshape]  {$\text{Concat}$};

 \node (attention_block) [above=3.95cm of background.south,rectangle, minimum width=10.cm, minimum height=9.15cm,  line width=1pt,draw=none, fill=attention_block,yshift=0cm]    {};

 \node (fully_connected_query) [process, xshift=3.1cm, yshift=0.9cm,  minimum width=3cm, minimum height=0.9cm] at (attention_block.south west) {\shortstack{$\text{FC,}\mathbf{W}^{\mathrm{(Q_{\ell}|\mathbf{K}^\mathbf{(c)}})}$}};
 \node (fully_connected_key) [process, xshift=-3.1cm, yshift=0.9cm,  minimum width=3.5cm, minimum height=0.9cm] at (attention_block.south east)  {\shortstack{$\text{FC,}\mathbf{W}^{\mathrm{(K^{\mathbf{(c)}})}}$}};
 \node (query) [below of = fully_connected_query,xshift = 0cm,yshift=0cm]  {};
\node (way)[waypoint] [xshift=0.476cm,yshift=-0.255cm] at ($(class_tokens)!0.49!(fully_connected_key)$) {};
\node (way2)[waypoint] [xshift=-0.61cm,yshift=-0.837cm] at ($(way)!0.38!(fully_connected_key)$) {};
\node (transpose) [process, above of = fully_connected_key,xshift=0.cm, yshift=0.5cm,  minimum width=1.5cm, minimum height=0.9cm]  {\shortstack{$\text{()}^{\mathsf{T}}$}};
 \node (label) [minimum width=3cm,xshift=-1.55cm,yshift=0.18cm, minimum height=0.9cm,font=\Large] at (query)  {${(T + S) \times d}$};
\node (cross) [draw, circle, minimum size=0.2cm,yshift=0.39cm,fill=conv,above of=fully_connected_query, line width=1] {\CircularCross};
\node (label3) [minimum width=3cm,xshift=2.4cm,yshift=0.0cm, minimum height=0.9cm,font=\Large,left of=cross]   {${\frac{1}{\sqrt{d}}}$};
\node (matrix_mul) [process, xshift=0cm, yshift=0.25cm,  minimum width=3cm, minimum height=0.9cm] at (attention_block.center)  {\shortstack{$\text{Matrix Multiplication}$}};
\node (z) [ xshift=0.5cm, yshift=-0.56cm,  above of=matrix_mul]   {\shortstack{$\mathbf{Z}$}};
\node (drop_op) [process, xshift=0cm, yshift=0.3cm,  minimum width=3cm, minimum height=0.9cm,above of=matrix_mul]   {\shortstack{$\text{Drop Operation}$}};
 \node (softmax) [process, xshift=0cm, yshift=0.3cm,  minimum width=3cm, minimum height=0.9cm,above of=drop_op]   {\shortstack{$\text{Softmax}$}};
\node (g) [ xshift=0.5cm, yshift=-0.48cm,  above of=softmax]   {\shortstack{$\mathbf{G}$}};
\node (v_k) [ xshift=1.6cm, yshift=0.07cm,  right of=g]   {\shortstack{$\mathbf{V}^{(\mathrm{c})}=\mathbf{K}^{(\mathrm{c})}$}};
\node (matrix_mul2) [process, xshift=0cm, yshift=0.45cm,  minimum width=3cm, minimum height=0.9cm,above of=softmax]   {\shortstack{$\text{Matrix Multiplication}$}};
\node (reshape2) [process, xshift=0cm, yshift=0.5cm,  minimum width=3cm, minimum height=0.9cm,above of=matrix_mul2]   {\shortstack{$\text{Reshape}$}};
 \node (H) [ xshift=0.5cm, yshift=-0.28cm,  above of=reshape2]   {\shortstack{$\mathbf{H}$}};
 \node (label3) [minimum width=3cm,xshift=-2cm,yshift=0.0cm, minimum height=0.9cm,font=\Large,left of=softmax]  {\shortstack{$\textbf{Attention}$ \\ $\textbf{Weights}$}};
 
  \node (cross_attention_label) [minimum width=2cm, minimum height=2cm, xshift=-2.35cm, yshift=-0.8cm, right=of reshape2.north east,    align=center,         
    inner sep=2pt,         ] {\shortstack{$\mathbf{Single}$$\text{-}$$\mathbf{head}$ \\ $\mathbf{Custom}$\\ $\mathbf{Cross} \ $$\mathbf{attention}\quad$}};
 
\draw [arrow] (feature_tokens) to  node[midway, above, align=center,xshift=-1.39cm,yshift=-0.6cm,font=\Large] {${d \times \frac{H}{k} \times \frac{W}{k} }$}    (reshape);
\draw[arrow] (way2) to (fully_connected_key);

\draw[arrow] (class_tokens) to  node[midway, above, align=center,xshift=-1.1cm,yshift=-0.35cm,font=\Large] {${S \times d}$}  ++(0,1.64)  --  (concat.east);
\draw[arrow] (reshape.north) to  node[midway, above, align=center,xshift=-0.87cm,yshift=0.2cm,font=\Large] {${T \times d}$}  ++(0,0)  --  ([xshift=1.51cm]concat);

\draw[arrow] (concat) to  node[midway, above, align=center,xshift=1.28cm,yshift=-0.19cm] {$\mathbf{Q}_{\ell}|\mathbf{K}^\mathbf{(c)}$}   ++(0,0.8)  -- ++(0,0) --  (fully_connected_query);
 \draw [arrow] (fully_connected_key) to  node[midway, xshift=-0.87cm, yshift=0cm,font=\Large]  {${S \times d}$} (transpose);
\draw [arrow] (fully_connected_query) to  node[midway, xshift=-1.55cm, yshift=0cm,font=\Large]  {${(T + S) \times d}$} (cross);
\draw [arrow] (transpose) to  node[midway, xshift=-0.85cm, yshift=0cm,font=\Large]  {${d \times  S}$} ([xshift=-0.87cm] matrix_mul.south east);
\draw [arrow] (cross) to  node[midway, xshift=-1.55cm, yshift=0.cm,font=\Large]  {${(T + S) \times d}$} ([xshift=0.87cm] matrix_mul.south west);
\draw [arrow] (matrix_mul) to  node[midway, xshift=-1.58cm, yshift=0cm,font=\Large]  {${(T + S) \times S}$} (drop_op);
\draw [arrow] (drop_op) to  node[midway, xshift=-0.87cm, yshift=0cm,font=\Large]  {${T  \times S}$} (softmax);
\draw [arrow] (softmax) to  node[midway, xshift=-0.87cm, yshift=0cm,font=\Large]  {${T  \times S}$} (matrix_mul2);
\draw[arrow] (way) to  node[midway, above, align=center,xshift=-0.8cm,yshift=-1.75cm] {} ++(2.3,0) --++(0,11.721)  --  (matrix_mul2.east);
\draw [arrow] (matrix_mul2) to  node[midway, xshift=-0.87cm, yshift=0cm,font=\Large]  {${T  \times d}$} (reshape2);

\draw[arrow] (reshape2) -- ++(0,1.64) node[midway, xshift=-1.45cm, yshift=0.22cm,font = \Large] {${d \times \frac{H}{k} \times \frac{W}{k} }$};

\end{tikzpicture}}
    \caption{\textbf{Single-head custom cross attention} block utilized in Fig.\ \ref{fig:context_mining}. Here, $\mathbf{c}=\mathbf{K}^{(\mathbf{c})} \in \mathbb{R}^{S \times d}$ denotes the learnable class tokens, which are used to inject global class-level information into $\mathbf{f}_{\ell}$.}
    \label{fig:custom}
\end{figure}
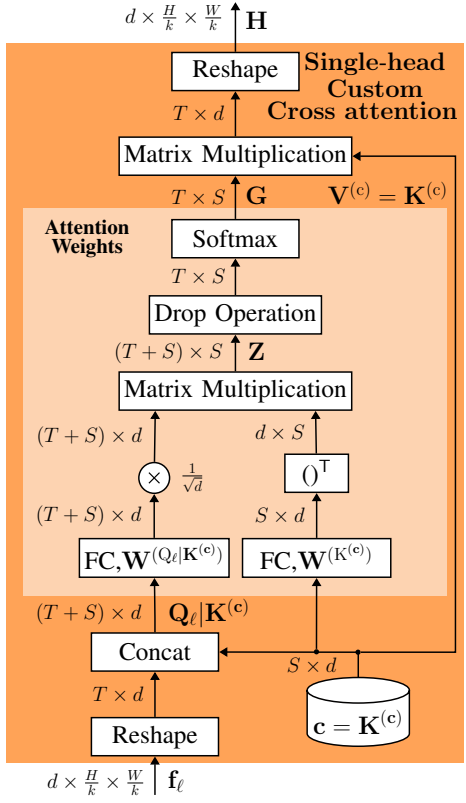

Both $\mathbf{D}$ and $\mathbf{JD}$ utilize a context mining block \cite{segdeformer}, as shown in Figure \ref{fig:context_mining}. The context mining block consists of two-stage attention blocks to refine $\bf{f}_{\ell}$. In the first stage, a single-head self attention block is applied to $\mathbf{f}_{\ell}$ with query, key, and value $\mathbf{Q}_{{\ell}}$ = $\mathbf{K}_{{\ell}}$ = $\mathbf{V}_{\ell} \in \mathbb{R}^{d \times \frac{H}{k} \times \frac{W}{k}}$. In parallel with the single-head attention block, there is a convolutional layer denoted as $\mathrm{Conv(\textit{h} \times \textit{h},\textit{F})}$, where $\textit{h} \times \textit{h}$ is the kernel size and $F$ is the number of filters, as well as a custom single-head cross attention block. The task of the single-head custom cross attention block is to inject class-level information to $\bf{f}_{\ell}$ through learned class tokens. The detailed architectures of the single-head attention and custom cross attention blocks are illustrated in Figures \ref{fig:naive} and \ref{fig:custom}, respectively. In the second stage, the output of the single-head custom cross attention block becomes the key and value $\mathbf{K}^{(\mathbf{H})} = \mathbf{V}^{(\mathbf{H})} \in \mathbb{R}^{d \times \frac{H}{k} \times \frac{W}{k}}$, while the query $\mathbf{Q}_{\ell}$ serves as input to the final single-head attention block. The outputs of the first-stage convolutional layer, the single-head attention block, and the second-stage single-head attention block are added to produce context mining block output $\mathbf{N} \in \mathbb{R}^{d \times \frac{H}{k} \times \frac{W}{k}}$.

Although the multi-stage attention blocks in the context mining block effectively capture complex local and global semantic relationships \cite{segdeformer}, they incur a high computational complexity as the size of $\mathbf{f}_{\ell}$ increases. \textit{To address this limitation, the proposed transformer-based joint feature and task decoder $\mathbf{JD}$ operates directly on $\mathbf{r}$ and $\mathbf{\hat{r}}$ for respective application}, both of which have a smaller number of feature channels ($d = 48$ vs.\ $d = 256$) and lower spatial resolution ($k = 8$ vs.\ $k = 4$), \textit{thereby significantly reducing the computational complexity of the context mining block}.

For the $\mathbf{JD}$, the spatial resolution of $\mathbf{N}$ is downsampled by a factor of $k=8$, which limits its ability to recover fine spatial details without intermediate upsampling. \textit{To address this, we let $\mathbf{N}$ pass through a light-weight grouped transposed convolution} $\mathrm{DWUpConv(\textit{h} \times \textit{h},\textit{F},\textit{G},\rho=2)}$ \cite{depthwiseseparable}, where $\mathrm{\textit{h} \times \textit{h}}$ is the kernel size, $F$ is the number of kernels, $G$ is the number of groups present in the layer, and stride $\rho=2$. After that, it is passed through another convolutional layer producing $S$ output feature maps, where $S$ represent the number of classes. This is followed by an upsampling layer, which is represented by $\mathrm{UpSampling}(u \times v)$, where $u \times v$ is the desired spatial size of the input features. This results in the network prediction $\bm{y}=(y_{i,s})$$\ \in \mathbb{I}$$^{S \times H \times W}$ with classes $\mathcal{S}=\{ 1,2, .., S \}$, class index $s \in \mathcal{S}$, pixel index $i \in \mathcal{I}$, and pixel index set $\mathcal{I} \in \{1,.., H \cdot W\}$, which is used to generate the semantic segmentation map as $\mathbf{m}=(m_{i})$ with $m_{i}=\mathrm{arg \ max}_{s \in \mathcal{S}} \ y_{i,s}$.

In Figure \ref{fig:naive}, the network architecture of the single-head self attention block utilized in Figure \ref{fig:overview_architectures} is illustrated. It takes first stage query $\mathbf{Q} = \mathbf{Q}_{\ell} \in \mathbb{R}^{d \times \frac{H}{k} \times \frac{W}{k}}$, key $\mathbf{K} \in \mathbb{R}^{d \times \frac{H}{k} \times \frac{W}{k}}$, and value $\mathbf{V} \in \mathbb{R}^{d \times \frac{H}{k} \times \frac{W}{k}}$ as input. In the first stage $\mathbf{K}=\mathbf{K}_{\ell}$, $\mathbf{V}=\mathbf{V}_{\ell}$, whereas in the second stage, it is $\mathbf{K}=\mathbf{K}^{(\mathbf{H})}$, $\mathbf{V}=\mathbf{V}^{(\mathbf{H})}$. The single-head self attention block contains convolutional projection layers, which are represented by $\mathrm{Conv(\textit{h} \times \textit{h},\textit{F})}$, where $\mathrm{\textit{h} \times \textit{h}}$ is the kernel size and $F$ is the number of kernels. Each of $\mathbf{Q}_{\ell}$, $\mathbf{K}$, and $\mathbf{V}$ is projected through separate projection layers and reshaped to $\mathbf{Q}_{\ell}^{\prime} \in \mathbb{R}^{T \times d}$, $\mathbf{K}^{\prime} \in \mathbb{R}^{T \times d}$ and $\mathbf{V}^{\prime} \in \mathbb{R}^{T \times d}$, where $\textit{T} = \frac{H}{k} \cdot \frac{W}{k} \in \mathbb{N}$ represents the (integer) number of input feature tokens. The reshaped $\mathbf{K}^{\prime}$ is transposed using $()^{\mathsf{T}}$ and multiplied with the projected $\mathbf{Q}_{\ell}$. The result of this multiplication is passed through a $\mathrm{Softmax}$ operator to generate the attention weights $\mathbf{G(Q^{\prime}_{\mathrm{\ell}},K^{\prime})} \in \mathbb{R}^{T \times T}$, which are then multiplied with $\mathbf{V^{\prime}} \in \mathbb{R}^{T \times d}$, and the output is reshaped and passed to the last projection layer to produce single-head attention block output $\mathbf{H} \in \mathbb{R}^{d \times \frac{H}{k} \times \frac{W}{k}}$.

In Figure \ref{fig:custom}, the network architecture of the single-head custom cross attention block utilized in Figure \ref{fig:context_mining} is depicted. This block takes features $\mathbf{f}_{\ell} \in \mathbb{R}^{d \times \frac{H}{k} \times \frac{W}{k}}$ as input, which are reshaped to $\mathbf{f}_{\ell} \in \mathbb{R}^{T \times d}$. Besides input features $\mathbf{f}_{\ell}$, it also utilizes learnable class tokens $\mathbf{c} \in \mathbb{R}^{S \times d}$ with classes $\mathcal{S} = \{1, 2, \ldots, S\}$, class index $s \in \mathcal{S}$, and number of classes $S$. The class tokens $\mathbf{c}=\mathbf{K}^{(\mathbf{c})} \in \mathbb{R}^{S \times d}$ are randomly initialized as trainable parameters to inject global class-level information into $\bf{f}_{\ell}$. After reshaping $\mathbf{f}_{\ell}$, a concatenation operation is performed between $\bf{f}_{\ell}$ and $\mathbf{c}=\mathbf{K}^{(\mathbf{c})} \in \mathbb{R}^{S \times d}$ to generate the query $(\mathbf{Q}_{\ell}{|\mathbf{K}^{(\mathbf{c})}}) \in \mathbb{R}^{(T+S) \times d}$. Unlike typical cross attention blocks, the value $\mathbf{V}^{(\mathbf{c})}=\mathbf{K}^{(\mathbf{c})}$ is \textit{not} subject to projection. For the query and the key, fully connected (FC) projection layers are employed with weights $\mathbf{W}^{\mathrm{(\mathbf{Q}_{\ell}{|\mathbf{K}^{(\mathbf{c})}})}},\mathbf{W}^{\mathrm{(K^{(\mathbf{c})})}} \in \mathbb{R}^{d \times d}$, to compute

\begin{equation}
    \mathbf{Z} = \frac{1}{\sqrt{d}}\mathbf{W}^{\mathrm{(\mathbf{Q}_{\ell}{|\mathbf{K}^{(\mathbf{c})}})}}\cdot{\mathrm{(\mathbf{Q}_{\ell}{|\mathbf{K}^{(\mathbf{c})}})}}\cdot\big(\mathbf{W}^{\mathbf{K}^{(\mathbf{c})}}\cdot\mathbf{K}^{(\mathbf{c})}\big)^{\mathsf{T}}.
    \label{Eq:Z}
\end{equation}

\noindent After projecting query and key and matrix multiplication~(\ref{Eq:Z}), $\mathbf{Z} \in \mathbb{R}^{(T+S) \times S}$ is passed through a $\mathrm{drop \ operation}$. The $\mathrm{drop \ operation}$ discards the class tokens and is defined as $\mathcal{F}: \mathbb{R}^{(T+S) \times S} \xrightarrow{} \mathbb{R}^{T \times S}$ with $\mathcal{F}(\mathbf{Z})=\mathbf{Z}_{1:T,:}$. Its output is passed to the $\mathrm{Softmax}$ operator to generate the attention weights $\mathbf{G}(\mathbf{Z}) \in \mathbb{R}^{T \times S}$, which are used to compute custom single-head cross attention output $\mathbf{H}= \mathrm{Reshape(\mathbf{G}\cdot\mathbf{K}^{(\mathbf{c})})}  \in \mathbb{R}^{d \times \frac{H}{4} \times \frac{W}{4}}$, where $\mathrm{Reshape}: \mathbb{R}^{T \times d} \xrightarrow{} \mathbb{R}^{d \times \frac{H}{4} \times \frac{W}{4}}$.

\section{EXPERIMENTAL SETUP}
\begin{table}[t!]
  \caption{\textbf{Datasets \& splits} used in our experiments.  }
  \label{table:datasets}
  \centering
  \begin{tabular}{@{}cccc@{}}
    \toprule
     \centering Dataset &  Official subsets & \#images & Symbol \\
    \midrule
   \centering \multirow{2}{*}{ADE20K \cite{ADE20K} }   & train & 20,210 & $\mathcal{D}_{\mathrm{ADE20K}}^{\mathrm{train}}$ \\  [0.2ex]
   \centering    & val & 2,000 & $\mathcal{D}_{\mathrm{ADE20K}}^{\mathrm{val}}$ \\ 
     \midrule
\centering \multirow{2}{*}{Cityscapes \cite{cityscapes}}   & train & 2,975 & $\mathcal{D}_{\mathrm{CS}}^{\mathrm{train}}$ \\ 
  \centering    & val & 500 & $\mathcal{D}_{\mathrm{CS}}^{\mathrm{val}}$ \\
  \bottomrule
  \end{tabular}
\end{table}

\subsection{DATASETS}
In Table \ref{table:datasets}, the datasets and splits used in our experiments are listed. We provide results on both ADE20K and Cityscapes datasets. The ADE20K dataset \cite{ADE20K} presents a significant challenge due to its diverse object categories, densely annotated scenes, and large number of classes. In total, it has $150$ classes spanning across both indoor and outdoor environments. Furthermore, we also report results on Cityscapes \cite{cityscapes}, which focuses on urban scene understanding, especially in the context of autonomous driving. It contains high-resolution images with fine-grained pixel-level annotations across $19$ semantic classes. Both datasets enable us to evaluate the generalization ability of our proposed approach against the current SOTA semantic segmentation methods for both in-car and distributed applications.
\begin{table}[t]
  \caption{\textbf{Hyperparameters used for training} the proposed \textsf{in-car JD} and \textsf{distributed JD} on the ADE20K and Cityscapes datasets. Here, \texttt{main} and \texttt{aux} denote the hyperparameters used for \texttt{main optimizer} and \texttt{auxiliary optimizer}, respectively.}
  \centering
  \begin{tabular}{@{}lcc@{}}
    \toprule
     \centering \textbf{Hyperparameter} &  ADE20K & Cityscapes  \\
    \midrule
     $\#$ of training iterations  & $160,000$  & $160,000$  \\ 
     Batch size   & $16$ & $8$  \\ 
     Random crop  & $512 \times 512$ & $768 \times 768$   \\ 
     Initial learning rate encoder (\texttt{main})   & $6\cdot10^{-5}$ & $6\cdot10^{-5}$  \\ 
     Initial learning rate decoder (\texttt{main})   & $6\cdot10^{-4}$ & $6\cdot10^{-4}$  \\ 
     Initial learning rate (\texttt{aux})  & $1\cdot10^{-3}$ & $1\cdot10^{-3}$  \\ 
     Learning rate schedule (\texttt{main} and \texttt{aux})  & polynomial & polynomial \\ 
     Optimizer (\texttt{main}) & AdamW & AdamW \\
     Optimizer (\texttt{aux}) & Adam & Adam \\
     Clip grad type (\texttt{aux})   & norm  & norm  \\ 
     Clip grad value (\texttt{aux})   & $1.0$  & $1.0$  \\ 
     Optimizer parameters $\beta_{1}$, $\beta_{2}$ (\texttt{main})   & $0.9, 0.999$ & $0.9, 0.999$  \\ 
     Weight decay (\texttt{main})   & $0.01$ & $0.01$  \\ 
  \bottomrule
  \end{tabular}
\label{table:hyperparameters}
\end{table}

 \begin{table*}[t!]
  \caption{\textbf{In-car application: Comparison of FLOPs, $\#$params, fps, and mIoU} measured with an input resolution of $2048 \times 1024$ for $\mathcal{D}_{\mathrm{CS}}^{\mathrm{val}}$ and $512 \times 512$ for $\mathcal{D}_{\mathrm{ADE20K}}^{\mathrm{val}}$ for our proposed transformer-based {\normalfont \textsf{in-car JD}} vs.\ two baselines.} 
  \label{table:in_car_results}
  \centering
  \setlength{\tabcolsep}{4pt} 
  \begin{tabular}{@{}llrrrrrrrr@{}}
    \toprule
    \multirow{2}{*}{\textbf{Encoder}} & \multirow{2}{*}{\textbf{Methods}} 
    & \multicolumn{4}{c}{\textbf{ADE20K} $\mathcal{D}_{\mathrm{ADE20K}}^{\mathrm{val}}$} 
    & \multicolumn{4}{c}{\textbf{Cityscapes} $\mathcal{D}_{\mathrm{CS}}^{\mathrm{val}}$} \\
    \cmidrule(lr){3-6} \cmidrule(lr){7-10}
    & & \multicolumn{1}{c}{\makecell{FLOPs \\[-2pt] \hspace{2pt}(G)}} & \multicolumn{1}{c}{\makecell{$\#$params\\[-2pt] \hspace{4pt}(M)}} & \multicolumn{1}{c}{fps} & \multicolumn{1}{c}{\makecell{mIoU \\[-2pt] \hspace{1pt}(\%)}} 
      & \multicolumn{1}{c}{\makecell{FLOPs \\[-2pt] \hspace{2pt}(G)}} & \multicolumn{1}{c}{\makecell{$\#$params \\[-2pt] \hspace{4pt}(M)}} & \multicolumn{1}{c}{fps} & \multicolumn{1}{c}{\makecell{mIoU \\[-2pt] \hspace{1pt}(\%)}} \\
   
    \midrule
        \multirow{2}{*}{\texttt{ResNet-50}} & \textsf{In-Car Baseline}  \cite{ahuja2023neural}         & 269 & 65.79 & 128.8 & \textbf{42.30}\textsuperscript{\scriptsize$\pm$ 0.8} & 2,149 & 65.72 & 7.8 & \textbf{78.72}\textsuperscript{\scriptsize$\pm$ 0.9} \\
                                     & \textsf{In-Car JD Baseline} \cite{nazirjd}  & \textbf{105} &\textbf{28.38} & \textbf{183.8} & 36.73\textsuperscript{\scriptsize$\pm$ 0.3} & \textbf{838} & \textbf{28.31} & \textbf{14.3} & 78.09\textsuperscript{\scriptsize$\pm$ 0.5} \\
    \midrule
    \midrule
    
    \multirow{2}{*}{\texttt{MiT-B0}} & \textsf{In-Car Baseline} \cite{segdeformer}         & 443 & 5.08 & 43.3 & \textbf{37.80}\textsuperscript{\scriptsize$\pm$ 0.2} & 17,261 & 4.98 & 1.4 & 76.73\textsuperscript{\scriptsize$\pm$ 3.7} \\
                                     & \textsf{In-Car JD (ours)}  & \textbf{14} & \textbf{3.80} & \textbf{154.3} & 37.20\textsuperscript{\scriptsize$\pm$ 0.4} & \textbf{408} & \textbf{3.78} & \textbf{16.5} & \textbf{76.93}\textsuperscript{\scriptsize$\pm$ 3.2} \\
    \midrule
    \multirow{2}{*}{\texttt{MiT-B1}} & \textsf{In-Car Baseline} \cite{segdeformer}         & 452 & 15.15 & 44.1 & 41.53\textsuperscript{\scriptsize$\pm$ 2.2} & 17,462 & 14.94 & 1.4 & 78.86\textsuperscript{\scriptsize$\pm$ 1.1} \\
                                     & \textsf{In-Car JD (ours)}  & \textbf{23} & \textbf{13.76} & \textbf{150.0} & \textbf{41.76}\textsuperscript{\scriptsize$\pm$ 0.5} & \textbf{609} & \textbf{13.74} & \textbf{14.7} & \textbf{79.03}\textsuperscript{\scriptsize$\pm$ 0.2} \\
    \midrule
    \multirow{2}{*}{\texttt{MiT-B2}} & \textsf{In-Car Baseline} \cite{segdeformer}         & 463 & 26.08 & 37.6 & \textbf{45.97}\textsuperscript{\scriptsize$\pm$ 0.8} & 17,753 & 25.98 & 1.3 & 80.85\textsuperscript{\scriptsize$\pm$ 0.2} \\
                                     & \textsf{In-Car JD (ours)}  & \textbf{34} & \textbf{24.80} & \textbf{92.8} & 45.50\textsuperscript{\scriptsize$\pm$ 0.5} & \textbf{900} & \textbf{24.79} & \textbf{10.0} & \textbf{80.97}\textsuperscript{\scriptsize$\pm$ 0.2} \\
    \midrule
    \multirow{2}{*}{\texttt{MiT-B5}} & \textsf{In-Car Baseline} \cite{segdeformer}         & 519 & 83.45 & 21.9 & 49.31\textsuperscript{\scriptsize$\pm$ 1.1} & 19,034 & 83.23 & 1.1 & 81.80\textsuperscript{\scriptsize$\pm$ 1.6} \\
                                     & \textsf{In-Car JD (ours)}  & \textbf{90} & \textbf{82.05} & \textbf{34.8} & \textbf{49.41}\textsuperscript{\scriptsize$\pm$ 0.2} & \textbf{2,181} & \textbf{82.03} & \textbf{4.5} & \textbf{81.87}\textsuperscript{\scriptsize$\pm$ 0.7} \\
    \midrule
    \bottomrule
\end{tabular}

\end{table*}

\subsection{EXPERIMENTAL DESIGN, TRAINING, AND METRICS}

We used \texttt{MMSegmentation} toolbox \cite{mmseg2020} for all our in-car application experiments, and combined it with the \texttt{CompressAI} \cite{begaint2020compressai} library for the distributed application experiments. All of the hyperparameter details for reproducing our results are listed in Table \ref{table:hyperparameters}. For the distributed application, we employed two optimizers: a \texttt{main optimizer} and an \texttt{auxiliary optimizer} \cite{nazirjd,begaint2020compressai}. The \texttt{auxiliary optimizer} updates the learnable \texttt{quantiles} parameter of the \texttt{EntropyBottleneck}, which defines the discretized probability distribution used to model $\mathbf{\hat{r}}$ and $\mathbf{\hat{h}}$ for entropy coding \cite{begaint2020compressai}. All other learnable parameters are optimized using the \texttt{main optimizer}. For the in-car application, we only use the \texttt{main optimizer}. During training, we utilized random cropping with an input resolution of $512 \,\times\, 512$ for ADE20K and $768 \,\times\, 768$ for Cityscapes \cite{cityscapes}. Further, we employ a batch size of $16$ for ADE20K, whereas for Cityscapes, we used $8$. We follow a two-stage training strategy: In the first stage, we train for the distributed application employing the same procedure proposed by \cite{nazirjd}. In the second stage, we focus on the in-car application, where we fine-tune only $\mathbf{JD}$, while keeping the rest of the network weights frozen. All trainings and evaluations are conducted on an \texttt{NVIDIA H100} GPU. 

We employ a \textsf{no compression baseline}, without applying a source codec, to evaluate the rate–distortion (RD) performance of our proposed transformer-based \textsf{distributed JD}. For both \textsf{no compression baseline} and \textsf{in-car baseline}, we select the current state-of-the-art (SOTA) semantic segmentation task decoder $\mathbf{D}$ \texttt{SegDeformer} \cite{segdeformer}. Additionally, we extend the method from \cite{nazirjd} to create an \textsf{in-car JD baseline}. For the \textsf{distributed baseline} and \textsf{distributed JD baseline}, we select the approaches by \cite{ahuja2023neural} and \cite{nazirjd}, respectively. Both \cite{ahuja2023neural,nazirjd} utilize a convolutional neural network (CNN)-based encoder $\mathbf{E}$ and decoder $\mathbf{D}$. We select \texttt{ResNet-50}\cite{resnet50} as $\mathbf{E}$ with \texttt{DeepLabV3} \cite{deeplabv3} as $\mathbf{D}$. Further, to ensure a fair comparison, we extended the \textsf{distributed baseline} \cite{ahuja2023neural} from CNN-based to a transformer-based setting. Specifically, we employ \texttt{MiT-B0}, \texttt{MiT-B1}, \texttt{MiT-B2} and \texttt{MiT-B5} as $\mathbf{E}$ with \texttt{SegDeformer} $\mathbf{D}$ for Cityscapes and ADE20K.

For our proposed transformer-based setting, we utilize four different variants of $\mathbf{E}$ from \texttt{SegFormer} \cite{Segformer}: \texttt{MiT-B0}, \texttt{MiT-B1}, \texttt{MiT-B2}, and \texttt{MiT-B5}. Further, we employ the proposed joint feature and task decoder $\mathbf{JD}$ with internal dimension $d=48$, as shown in Fig.\ \ref{fig:overview_architectures}(b). The variants of $\mathbf{E}$ cover a wide range of model complexities, from lightweight (\texttt{MiT-B0}) to heavyweight (\texttt{MiT-B5}) and vary significantly in model capacity and computational cost. They enable us to explore the trade-offs between mIoU and the computational complexities for both in-car and distributed applications. Additionally, they allow us to evaluate the generalizability of our method across varying sizes of $\mathbf{E}$, ensuring that performance gains are not limited to a specific model size. 

To evaluate the performance of the in-car application, we use the mIoU metric, which is commonly used in semantic segmentation tasks. For the distributed application, we report the rate-distortion (RD) performance. The rate is defined following (\ref{Eq:rate}) as bits per pixel (bpp), and distortion is measured using the mIoU metric following (\ref{Eq:CE}). In addition, we compare the computational complexity of our approach against the prior methods by reporting the number of floating-point operations per image (FLOPs), number of parameters, and frames per second (fps). The FLOPs, fps, and mIoU are measured at the resolutions of $2048 \times 1024 \times 3$ and $512 \times 512 \times 3$ for Cityscapes and ADE20K, respectively. The fps were measured on an \texttt{NVIDIA H100} GPU.

 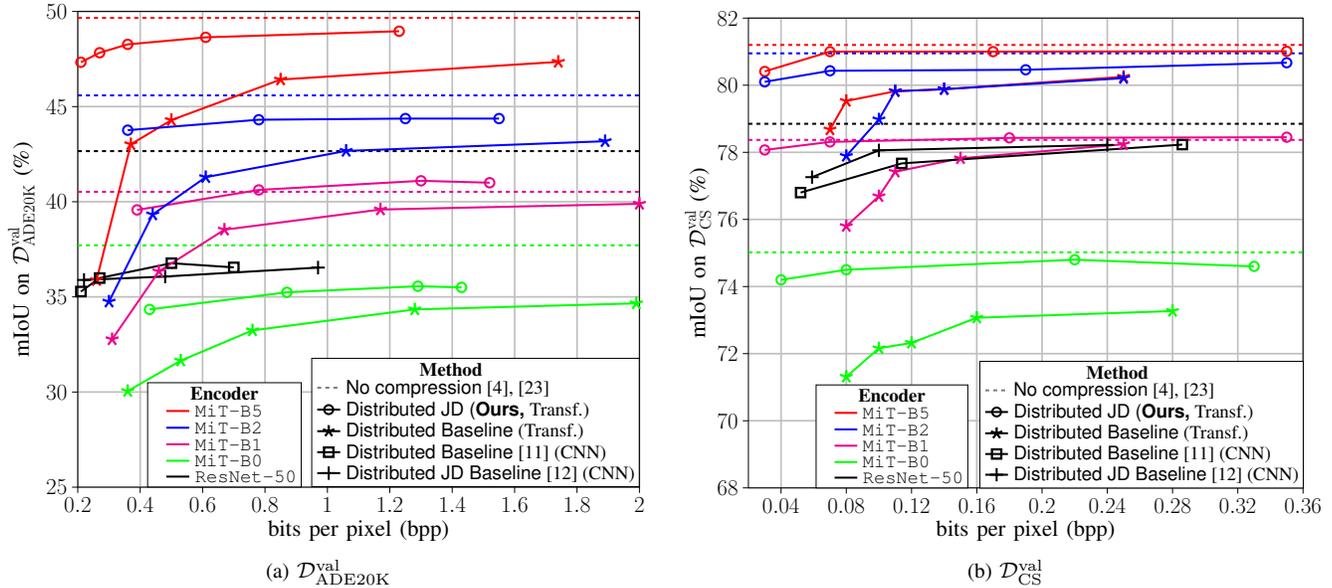
\begin{figure*}[t!]
 \hspace{-0.85em}
  \subfloat[$\mathcal{D}_{\mathrm{ADE20K}}^{\mathrm{val}}$\label{fig:ade20k_results}]{
    \resizebox{0.488\linewidth}{!}{


\begin{tikzpicture}

\definecolor{darkgoldenrod1811370}{RGB}{181,137,0}
\definecolor{darkgreen}{RGB}{0,100,0}
\definecolor{lightgray}{RGB}{211,211,211}
\definecolor{lightgray204}{RGB}{204,204,204}
\definecolor{rosybrown163154141}{RGB}{163,154,141}
\definecolor{voilet}{RGB}{139,69,19}
\begin{axis}[
width=16.25cm,   
legend cell align={left},
legend cell align={left},
legend cell align={left},
scaled ticks=false,
tick align=outside,
tick pos=left,
xtick distance=0.2,
tick label style={font=\LARGE},
label style={font=\LARGE},
xlabel={bits per pixel (bpp) },
xmajorgrids,
xmin=0.2, xmax=2,
xtick style={color=black},
x tick label style={/pgf/number format/precision=10},
x grid style={gray!50, line width=1.25pt},
y grid style={gray!50, line width=1.25pt},
ylabel={\(\displaystyle \textrm{mIoU on $\mathcal{D}_{\mathrm{ADE20K}}^{\mathrm{val}}$}\) (\%)},
ylabel style={yshift=7pt},
xlabel style={yshift=-6pt},
ymajorgrids,
ymin=25, ymax=50,
ytick style={color=black},
ytick={25,30,35,40,45,50},
yticklabels={
 \(\displaystyle {25}\),
  \(\displaystyle {30}\),
  \(\displaystyle {35}\),
  \(\displaystyle {40}\),
  \(\displaystyle {45}\),
  \(\displaystyle {50}\)
}
]
\addplot [line width=1.5pt, red, mark size=3pt, mark options={ line width=1.5, solid, fill=none}, dashed]
table {%
0.2 49.66
2 49.66
};
\label{No_Compression_mitb5_ade20k}

\addplot [line width=1.5pt,  blue,mark size=3pt, mark options={ line width=1.5, solid, fill=none}, dashed]
table {%
0.2 45.59
2 45.59
};
\label{No_Compression_mitb2_ade20k}

\addplot [line width=1.5pt, black,  mark size=3pt, mark options={ line width=1.5, solid, fill=none}, dashed]
table {%
0.2 42.66
2 42.66
};
\label{No_Compression_resnet50_ade20k}

\addplot [line width=1.5pt,  magenta, mark size=3pt, mark options={ line width=1.5, solid, fill=none}, dashed]
table {%
0.2 40.52
2 40.52
};
\label{No_Compression_mitb1_ade20k}

\addplot [line width=1.5pt,  green, mark size=3pt, mark options={ line width=1.5, solid, fill=none}, dashed]
table {%
0.2 37.71
2 37.71
};
\label{No_Compression_mitb0_ade20k}

\addplot [line width=1.5pt,red, mark=star, mark size=5,  mark options={ line width=1.7, solid}]
table {%
0.26 35.88
0.37 43.02
0.50 44.29
0.85 46.42
1.74 47.35
};
\label{Ahuja_mitb5_ade20k}
\addplot [line width=1.5pt,  red,  mark=o, mark size=3.5,  mark options={ line width=1.5, solid}]
table {%
0.21 47.33
0.27 47.83
0.36 48.27
0.61 48.64
1.23 48.96
}; 
\label{ours_mitb5_ade20k}

\addplot [line width=1.5pt,black, mark=square, mark size=3.5,  mark options={ line width=1.7, solid}]
table {%
0.21 35.28
0.27 35.99
0.50 36.77
0.7 36.55
};
\label{Ahuja_resnet50_ade20k}
\addplot [line width=1.5pt,   black,  mark=+, mark size=5,  mark options={ line width=1.5, solid}]
table {%
0.22 35.89
0.48 36.07
0.97 36.54
};
\label{ours_resnet50_ade20k}

\addplot [line width=1.5pt,magenta, mark=star, mark size=5,  mark options={ line width=1.7, solid}]
table {%
0.31 32.77
0.46 36.34
0.67 38.53
1.17 39.59
2 39.89
};
\label{Ahuja_mitb1_ade20k}
\addplot [line width=1.5pt,  magenta,  mark=o, mark size=3.5,  mark options={ line width=1.5, solid}]
table {%
0.39 39.57
0.78 40.62
1.3 41.1
1.52 41.0
}; 
\label{ours_mitb1_ade20k}
\addplot [line width=1.5pt,blue, mark=star, mark size=5,  mark options={ line width=1.7, solid}]
table {%
0.30 34.75
0.44 39.33
0.61 41.29
1.06 42.67
1.89 43.18
};
\label{Ahuja_mitb2_ade20k}
\addplot [line width=1.5pt,  blue,  mark=o, mark size=3.5,  mark options={ line width=1.5, solid}]
table {%
0.36 43.76
0.78 44.31
1.25 44.37
1.55 44.37
}; 
\label{ours_mitb2_ade20k}
\addplot [line width=1.5pt,green, mark=star, mark size=5,  mark options={ line width=1.7, solid}]
table {%
0.36 30.05
0.53 31.65
0.76 33.24
1.28 34.34
1.99 34.66
};
\label{Ahuja_mitb0_ade20k}
\addplot [line width=1.5pt,  green,  mark=o, mark size=3.5,  mark options={ line width=1.5, solid}]
table {%
0.43 34.34
0.87 35.24
1.29 35.56 
1.43 35.5
};
\label{ours_mitb0_ade20k}

\addplot [line width=1.5pt, gray , dashed]
table {%
0.0 44
};
\label{No_Compression_fake_legend_ade20k}
\addplot [line width=1.5pt,black,mark=star, mark size=5,  mark options={ line width=1.7, solid}]
table {%
0.0520000457763672 34.8000030517578
};
\label{Ahuja_fake_legend_ade20k}
\addplot [line width=1.5pt,black,mark=o, mark size=3.5,  mark options={ line width=1.5, solid}]
table {%
0.0520000457763672 34.8000030517578
};
\label{ours_fake_legend}

\addplot [line width=1.5pt,  red]
table {%
0.4 34 
};
\label{ours_fake_legend_ade20k}

\addplot [line width=1.5pt,  red]
table {%
0.4 34 
};
\label{mit-b5}
\addplot [line width=1.5pt,  blue]
table {%
0.4 34 
};
\label{mit-b2}

\addplot [line width=1.5pt,  magenta]
table {%
0.4 34 
};
\label{mit-b1}

\addplot [line width=1.5pt,  green]
table {%
0.4 34 
};
\label{mit-b0}

\addplot [line width=1.5pt,  black]
table {%
0.4 34 
};
\label{resnet50}

\end{axis}

\node [draw,fill=white,font=\Large] at (rel axis cs:0.597,0.136) {
  \shortstack[l]{
   \hspace*{27mm}\textbf{Method}\\
    \ref*{No_Compression_fake_legend_ade20k} \textsf{No compression} \cite{deeplabv3,segdeformer} \\
    \ref*{ours_fake_legend} \textsf{Distributed JD (\textbf{Ours, }}Transf.) \\
    \ref*{Ahuja_fake_legend_ade20k} \textsf{Distributed Baseline} (Transf.) \\
    \ref*{Ahuja_resnet50_ade20k} \textsf{Distributed Baseline} \cite{ahuja2023neural} (CNN) \\
    \ref*{ours_resnet50_ade20k} \textsf{Distributed JD Baseline} \cite{nazirjd} (CNN)
  }
};

\node [draw,fill=white,font = \Large] at (rel axis cs: 0.15,0.11){\shortstack[l]{
\hspace*{10mm}\textbf{Encoder}\\
 \hspace{0.7em}\ref*{mit-b5} \texttt{MiT-B5}  \\  
  \hspace{0.7em}\ref*{mit-b2} \texttt{MiT-B2} \\ 
  \hspace{0.7em}\ref*{mit-b1} \texttt{MiT-B1} \\
   \hspace{0.7em}\ref*{mit-b0} \texttt{MiT-B0} \\
    \hspace{0.7em}\ref*{resnet50} \texttt{ResNet-50} 
  }
 };
 \end{tikzpicture}


}
  }\hfill
  \subfloat[$\mathcal{D}_{\mathrm{CS}}^{\mathrm{val}}$\label{fig:cityscapes_results}]{
    \resizebox{0.488\linewidth}{!}{\begin{tikzpicture}

\definecolor{darkgoldenrod1811370}{RGB}{181,137,0}
\definecolor{darkgreen}{RGB}{0,100,0}
\definecolor{lightgray}{RGB}{211,211,211}
\definecolor{lightgray204}{RGB}{204,204,204}
\definecolor{rosybrown163154141}{RGB}{163,154,141}
\definecolor{voilet}{RGB}{139,69,19}
\begin{axis}[
width=16.25cm,   
legend cell align={left},
legend cell align={left},
legend cell align={left},
scaled ticks=false,
tick align=outside,
tick pos=left,
tick label style={font=\LARGE},
label style={font=\LARGE},
xlabel={bits per pixel (bpp) },
xmajorgrids,
xmin=0.02, xmax=0.36,
xtick distance=0.04,
xtick style={color=black},
x tick label style={/pgf/number format/precision=10},
x grid style={gray!50, line width=1.25pt},
y grid style={gray!50, line width=1.25pt},
ylabel={\(\displaystyle \textrm{mIoU on $\mathcal{D}_{\mathrm{CS}}^{\mathrm{val}}$}\) (\%)},
ylabel style={yshift=3pt},
xlabel style={yshift=-6pt},
ymajorgrids,
ymin=68, ymax=82,
ytick style={color=black},
ytick={68,70,72,74,76,78,80,82},
yticklabels={
  \(\displaystyle {68}\),
  \(\displaystyle {70}\),
  \(\displaystyle {72}\),
  \(\displaystyle {74}\),
  \(\displaystyle {76}\),
  \(\displaystyle {78}\),
  \(\displaystyle {80}\),
  \(\displaystyle {82}\),
}
]

\addplot [line width=1.5pt, black, dashed]
table {%
0.02 78.85
0.36 78.85
};
\label{No_Compression_resnet50_cs}

\addplot [line width=1.5pt,  magenta, dashed]
table {%
0.02 78.37
0.36 78.37
};
\label{No_Compression_mitb1_cs}
\addplot [line width=1.5pt,  blue, dashed]
table {%
0.02 80.95
0.36 80.95
};
\label{No_Compression_mitb2_cs}

\addplot [line width=1.5pt,  red, dashed]
table {%
0.02 81.2
0.36 81.2
};
\label{No_Compression_mitb5_cs}
    
\addplot [line width=1.5pt,  green, dashed]
table {%
0.02 75.02
0.36 75.02
};
\label{No_Compression_mitb0_cs}

\addplot [line width=1.5pt,red, mark=star, mark size=5,  mark options={ line width=1.7, solid}]
table {%
0.07 78.68 
0.08 79.53
0.11 79.82
0.14 79.88
0.25 80.25
};
\label{Ahuja_mitb5_cs}

\addplot [line width=1.5pt,  red,  mark=o, mark size=3.5,  mark options={ line width=1.5, solid}]
table {%
0.03 80.41
0.07 81
0.17 81
0.35 81.01
};
\label{ours_mitb5_cs}
\addplot [line width=1.5pt,green, mark=star, mark size=5,  mark options={ line width=1.7, solid}]
table {%
0.08 71.31
0.10 72.16
0.12 72.32
0.16 73.07
0.28 73.27
};
\label{Ahuja_mitb0_cs}
\addplot [line width=1.5pt,  green,  mark=o, mark size=3.5,  mark options={ line width=1.5, solid}]
table {%
0.04 74.2
0.08 74.5
0.22 74.8
0.33 74.6
};
\label{ours_mitb0_cs}
\addplot [line width=1.5pt,magenta, mark=star, mark size=5,  mark options={ line width=1.7, solid}]
table {%
0.08 75.8
0.10 76.69
0.11 77.42
0.15 77.82
0.25 78.24
};
\label{Ahuja_mitb1_cs}
\addplot [line width=1.5pt,  magenta,  mark=o, mark size=3.5,  mark options={ line width=1.5, solid}]
table {%
0.03 78.07
0.07 78.31
0.18 78.43
0.35 78.45
};
\label{ours_mitb1_cs}
\addplot [line width=1.5pt,blue, mark=star, mark size=5,  mark options={ line width=1.7, solid}]
table {%
0.08 77.89
0.10 78.98
0.11 79.82
0.14 79.88
0.25 80.21
};
\label{Ahuja_mitb2_cs}
\addplot [line width=1.5pt,  blue,  mark=o, mark size=3.5,  mark options={ line width=1.5, solid}]
table {%
0.03 80.1
0.07 80.43
0.19 80.46
0.35 80.67
};
\label{ours_mitb2_cs}
\addplot [line width=1.5pt,black, mark=square, mark size=3.5,  mark options={ line width=1.7, solid}]
table {%
0.0520000457763672 76.8000030517578
0.113999962806702 77.6699981689453
0.28600001335144 78.2300033569336
};
\label{Ahuja_resnet50_cs}
\addplot [line width=1.5pt,   black,  mark=+, mark size=5,  mark options={ line width=1.5, solid}]
table {%
0.059 77.26 
0.1 78.06
0.24 78.22

};
\label{ours_resnet50_cs}

\addplot [line width=1.5pt, gray, , dashed]
table {
0.0 78.85
};
\label{No_Compression_fake_legend_cs}

\addplot [line width=1.5pt,black,draw=none,mark=star, mark size=4.5,  mark options={ line width=1.7, solid}]
table {
0.0120000457763672 77.8000030517578
};
\label{Ahuja_fake_legend_cs}

\addplot [line width=1.5pt,black,mark=o, mark size=3,  mark options={ line width=1.5, solid}]
table {
0.0019 78.26 
};
\label{ours_fake_legend_cs}

\addplot [line width=1.5pt,  red]
table {%
0.4 34 
};
\label{ours_fake_legend_ade20k}

\addplot [line width=1.5pt,  red]
table {%
0.4 34 
};
\label{mit-b5-cs}
\addplot [line width=1.5pt,  blue]
table {%
0.4 34 
};
\label{mit-b2-cs}

\addplot [line width=1.5pt,  magenta]
table {%
0.4 34 
};
\label{mit-b1-cs}

\addplot [line width=1.5pt,  green]
table {%
0.4 34 
};
\label{mit-b0-cs}

\addplot [line width=1.5pt,  black]
table {%
0.4 34 
};
\label{resnet50-cs}

\end{axis}

\node [draw,fill=white,font=\Large] at (rel axis cs: 0.2,-2.319){\shortstack[l]{
\hspace*{10mm}\textbf{Encoder}\\
 \hspace{0.7em}\ref*{mit-b5} \texttt{MiT-B5}  \\  
  \hspace{0.7em}\ref*{mit-b2} \texttt{MiT-B2} \\ 
  \hspace{0.7em}\ref*{mit-b1} \texttt{MiT-B1} \\
   \hspace{0.7em}\ref*{mit-b0} \texttt{MiT-B0} \\
    \hspace{0.7em}\ref*{resnet50} \texttt{ResNet-50} 
  }
 };

\node [draw,fill=white,font = \Large] at (rel axis cs: 0.65,-2.294){  \shortstack[l]{
   \hspace*{27mm}\textbf{Method}\\
    \ref*{No_Compression_fake_legend_ade20k} \textsf{No compression} \cite{deeplabv3,segdeformer} \\
    \ref*{ours_fake_legend} \textsf{Distributed JD (\textbf{Ours,}} Transf.) \\
    \ref*{Ahuja_fake_legend_ade20k} \textsf{Distributed Baseline} (Transf.) \\
    \ref*{Ahuja_resnet50_ade20k} \textsf{Distributed Baseline} \cite{ahuja2023neural} (CNN) \\
    \ref*{ours_resnet50_ade20k} \textsf{Distributed JD Baseline} \cite{nazirjd} (CNN)
  }
 };

\end{tikzpicture}}
  }
  \caption{\textbf{Distribution application: mIoU ($\%$) vs.\ bitrate in bits per pixel (bpp)} for the proposed {\normalfont \textsf{distributed JD}} vs.\ various baselines for the (a) $\mathcal{D}_{\mathrm{ADE20K}}^{\mathrm{val}}$ and the (b) $\mathcal{D}_{\mathrm{CS}}^{\mathrm{val}}$ dataset.}
  \label{fig:result_comparison}
\end{figure*}

\begingroup
\setlength{\belowcaptionskip}{-1pt}
\begin{table}[t!]
  \caption{\textbf{Distributed application: Comparison of FLOPs and $\#$params at the cloud}  measured with an input resolution of $2048\!\times\!1024$ for $\mathcal{D}_{\mathrm{CS}}^{\mathrm{val}}$ and $512 \times 512$ for $\mathcal{D}_{\mathrm{ADE20K}}^{\mathrm{val}}$ for the proposed {\normalfont \textsf{distributed JD}} vs.\ various baselines.}.
  \label{table:flops_params_comparison}
  \centering
  \setlength{\tabcolsep}{2.25pt} 
  \begin{tabular} {llrrrr}
    \toprule
    \multirow{3}{*}{\makecell{\textbf{Distributed} \\ \textbf{methods}}} & \multirow{3}{*}{\makecell{\textbf{Model} \\ \textbf{Type}}} & \multicolumn{2}{c}{\textbf{ADE20K} $\mathcal{D}_{\mathrm{ADE20K}}^{\mathrm{val}}$} & \multicolumn{2}{c}{\textbf{Cityscapes} $\mathcal{D}_{\mathrm{CS}}^{\mathrm{val}}$} \\
    \cmidrule(lr){3-4} \cmidrule(lr){5-6}
     & & \makecell{FLOPs \\[-2pt] \hspace{2pt}(G)} & \makecell{$\#$params \\[-2pt] \hspace{6pt}(M)} & \makecell{FLOPs \\[-2pt] \hspace{2pt}(G)} & \makecell{$\#$params \\[-2pt] \hspace{6pt}(M)} \\  
    \midrule
   \textsf{Dist. Baseline}  \cite{ahuja2023neural} & CNN & \underline{134}  & 42.77 & 2,616  & 42.44  \\  
    \textsf{Dist. Baseline}  & Transf. & 434  & \underline{1.38} & 7,925  & \underline{1.32}  \\  
    \textsf{Dist. JD Baseline}  \cite{nazirjd} & CNN & \textbf{5}  & 4.75 & \textbf{88}  & 4.92  \\
    \textsf{Dist. JD (ours)} & Transf. & \textbf{5} & \textbf{0.06} & \underline{200} & \textbf{0.05} \\
    \bottomrule
  \end{tabular}
\end{table}
\endgroup

\section{EXPERIMENTAL RESULTS AND DISCUSSION}
\subsection{IN-CAR SEMANTIC SEGMENTATION}

In Table \ref{table:in_car_results}, we present a detailed comparison of FLOPs (G), parameter count (M), frames per second (fps) and mIoU on Cityscapes $\mathcal{D}_{\mathrm{CS}}^{\mathrm{val}}$ and ADE20K $\mathcal{D}_{\mathrm{ADE20K}}^{\mathrm{val}}$ datasets, for \textsf{in-car baseline}, \textsf{in-car JD baseline}, and for our proposed transformer-based \textsf{in-car JD}. Further, we repeat \textsf{in-car JD baseline} along with each setting of \textsf{in-car baseline} and \textsf{in-car JD} three times with different random seeds and report the mean mIoU with a $95 \%$ confidence interval. All fps measurements are also averaged over three runs following a warm-up period \cite{mmseg2020}. 

In the upper segment of Table \ref{table:in_car_results}, two \texttt{ResNet-50}-based baseline methods are reported. Note their computational complexity being in the range $100$ ... $270$ GFLOPs (on ADE20K) and $830$ ... $2,150$ GFLOPs (on Cityscapes). The fps are weakly related to the GFLOPs. Concerning mIoU, \textsf{In-Car Baseline} \cite{ahuja2023neural} performs significantly better on ADE20K, however, at the price of a much larger and slower model.

In the lower segment of Table \ref{table:in_car_results}, we report on the \texttt{MiT-Bn} encoders. We observe that there is no significant difference in mIoU between the \texttt{MiT-Bn} \textsf{In-Car Baselines} \cite{segdeformer} and our proposed \textsf{In-Car JD}. The \textsf{In-Car Baseline}  \cite{segdeformer} delivers the lowest fps performance of all methods on both datasets in the table. 
The reason is that a standard attention mechanism is employed in \texttt{SegDeformer} \cite{segdeformer}. Our work addresses this poor fps performance so that our \textsf{In-Car JD} achieves a strong $34.8$ ... $154.3$ fps on ADE20K (\textsf{In-Car Baseline} reaches $21.9$ ... $44.1$ only), whereby we improve fps by a factor of up to $3.5$. On Cityscapes, our \textsf{In-Car JD} achieves $4.5$ ... $16.5$ fps (\textsf{In-Car Baseline} reaches a poor $1.1$ ... $1.4$ fps), whereby we improve fps by a factor of up to $11.7$. 

At the same time, our \textsf{In-Car JD} comes with a somewhat lower footprint than \textsf{In-Car Baseline} \cite{segdeformer}, and with a drastically reduced computational complexity that is even below the two \texttt{ResNet-50}-based baselines on ADE20K, and, at a comparable mIoU (i.e., \texttt{MiT-B1}) also on Cityscapes ($609$ GFLOPs vs.\ $838$ or $2,149$ GFLOPs). \textit{Accordingly, with our novel \textsf{In-Car JD} approach, we excel the transformer baseline \cite{segdeformer} in all metrics, and the \texttt{ResNet-50}-based baselines \cite{ahuja2023neural,nazirjd} at similar mIoU in all metrics on both datasets.}

\subsection{DISTRIBUTED SEMANTIC SEGMENTATION}

In Figure \ref{fig:result_comparison}, we aim at the distributed application and show the rate-distortion (RD) performance of our proposed transformer-based \textsf{distributed JD} against three baselines: the \textsf{distributed baseline} \cite{ahuja2023neural} (CNN), a transformer-based variant of \textsf{distributed baseline}, and the \textsf{distributed JD baseline} \cite{nazirjd} (CNN). Fig.\ \ref{fig:result_comparison}\subref{fig:ade20k_results} shows the results on ADE20K, while Fig.\ \ref{fig:result_comparison}\subref{fig:cityscapes_results} reports the results on Cityscapes, respectively. All of the results presented in Figure \ref{fig:result_comparison} utilize exactly the same source codec as described in Section \ref{sec:efficient_source_codec}.

On both datasets, we observe a mostly monotonic relation between bitrate and mIoU. Expectedly, larger encoders deliver better mIoU performance. \textit{Importantly, we observe that our proposed \textsf{Distributed JD} (circle markers) exceeds the performance of the \textsf{Distributed Baseline} ($\ast$ marker) at all bitrates.} For \texttt{MiT-B1}, at medium to high bitrates, we even observe an equal or better mIoU of our \textsf{Distributed JD} when compared to the uncompressed baseline \cite{segdeformer,deeplabv3} (dashed mIoU level). The comparative discussion of the \texttt{ResNet-50}-based approaches (black curves) to the \texttt{MiT-Bn}-based ones is best done in Table \ref{table:flops_params_comparison}, as model sizes and complexities are not transparent in Figure \ref{fig:result_comparison}.

In Table \ref{table:flops_params_comparison}, we show a comparison of FLOPs (G) and parameter count (M) at the cloud on ADE20K $\mathcal{D}_{\mathrm{ADE20K}}^{\mathrm{val}}$ and Cityscapes $\mathcal{D}_{\mathrm{CS}}^{\mathrm{val}}$ datasets for the proposed \textsf{JD} vs.\ the \textsf{distributed baseline} \cite{ahuja2023neural}, its transformer-based variant, and the \textsf{distributed JD baseline} \cite{nazirjd}. 
For our proposed $\mathbf{JD}$ and \textsf{distributed JD baseline}, these metrics refer to $\mathbf{CD}$ and $\mathbf{JD}$, while for the other baselines to $\mathbf{CD}$, $\mathbf{FD}$ and $\mathbf{D}$, cf. Fig.\ \ref{fig:high_level}. These metrics are critical for the scalability at the cloud in distributed applications \cite{nazirjd}. 

As displayed in Table \ref{table:flops_params_comparison}, our \textsf{Distributed JD} approach shows low computational 
complexity, much lower than the baselines without a $\mathbf{JD}$. The CNN-based \textsf{Distributed JD Baseline} has equal complexity on ADE20K, but even lower on Cityscapes, however, at the price of an up to about $100$-times larger cloud-based decoder. \textit{Concerning model size, our proposed \textsf{Distributed JD} is best in class using only $0.14$\% ($0.04$\%), i.e., $0.06$ M vs.\ $1.38$ M parameters on ADE20K ($0.05$ M vs.\ $1.32$ M parameters on Cityscapes)}. 

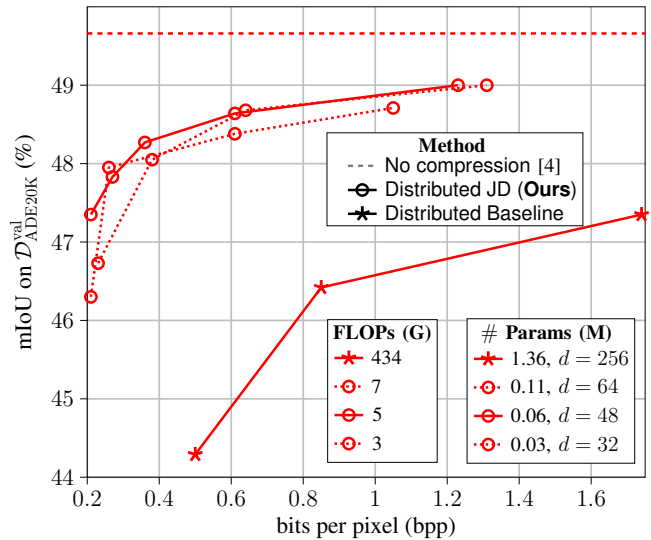
\begin{figure}[t!]
 \hspace{-0.4em}
    \resizebox{1\linewidth}{!}{
\begin{tikzpicture}

\definecolor{brightred}{RGB}{238, 75,43}
\definecolor{redbrown}{RGB}{165, 42, 42}
\definecolor{rubyred}{RGB}{224, 17, 95}
\definecolor{darkgoldenrod1811370}{RGB}{181,137,0}
\definecolor{darkgreen}{RGB}{0,100,0}
\definecolor{lightgray}{RGB}{211,211,211}
\definecolor{lightgray204}{RGB}{204,204,204}
\definecolor{rosybrown163154141}{RGB}{163,154,141}
\definecolor{voilet}{RGB}{139,69,19}
\begin{axis}[
width=13.25cm,   
legend cell align={left},
legend cell align={left},
legend cell align={left},
legend style={
  fill opacity=1.0,
  draw opacity=1,
  text opacity=1,
  at={(0.5,0.02)},
  anchor=south west,
  draw=white!80!black
},
legend style={
  fill opacity=1.0,
  draw opacity=1,
  text opacity=1,
  at={(0.7,0.02)},
  anchor=south west,
  draw=white!80!black
},
legend style={
  fill opacity=1.0,
  draw opacity=1,
  text opacity=1,
  at={(0.7,0.02)},
  anchor=south west,
  draw=white!80!black
},
tick align=outside,
tick pos=left,
tick label style={font=\Large},
label style={font=\Large},
xlabel={bits per pixel (bpp) },
xmajorgrids,
xmin=0.2, xmax=1.75,
xtick style={color=black},
x grid style={gray!50, line width=1pt},
y grid style={gray!50, line width=1pt},
ylabel={\(\displaystyle \textrm{mIoU on $\mathcal{D}_{\mathrm{ADE20K}}^{\mathrm{val}}$}\) (\%)},
ylabel style={yshift=3pt},
xlabel style={yshift=-5pt},
ymajorgrids,
ymin=44, ymax=50,
ytick style={color=black},
ytick={44,45,46,47,48,49},
yticklabels={
 \(\displaystyle {44}\),
 \(\displaystyle {45}\),
 \(\displaystyle {46}\),
 \(\displaystyle {47}\),
 \(\displaystyle {48}\),
 \(\displaystyle {49}\)
}
]
\addplot [line width=1.5pt,  red, dashed]
table {%
0.2 49.66
1.75 49.66
};
\label{No_Compression_ablation_mitb0_ade20k}
-
\addplot [line width=1.5pt,red,mark=star, mark size=4.5,  mark options={ line width=1.7, solid}]
table {%
0.5 44.29
0.85 46.42
1.74 47.35
};
\label{Ahuja_ablation_mitb0_ade20k}
\addplot [line width=1.5pt, solid, red,  mark=o, mark size=3.35,  mark options={ line width=1.5, solid}]
table {%
0.21 47.35
0.27 47.83
0.36 48.27
0.61 48.64
1.23 49
};
\label{ours_mitb0_48_ade20k}
\addplot [line width=1.5pt,  red, dotted, mark=o, mark size=3.35,  mark options={ line width=1.5, solid}]
table {
0.23 46.73
0.38 48.05
0.64 48.68
1.31 49.0
};
\label{ours_mitb0_ade20k}
\addplot [line width=1.5pt, dotted, red,  mark=o, mark size=3.35,  mark options={ line width=1.5, solid}]
table {%
0.21 46.30
0.26 47.95
0.61 48.38
1.05 48.71
};
\label{ours_mitb0_32_ade20k}

\addplot [line width=1.5pt, gray , dashed]
table {%
0.0 44
};
\label{No_Compression_fake_legend_ade20k}
\addplot [line width=1.5pt,black,mark=star, mark size=4.5,  mark options={ line width=1.7, solid}]
table {%
0.0520000457763672 34.8000030517578
};
\label{Ahuja_fake_legend_ade20k}
\addplot [line width=1.5pt,black,mark=o, mark size=3.35,  mark options={ line width=1.5, solid}]
table {%
0.0520000457763672 34.8000030517578
};
\label{ours_fake_legend}

\end{axis}

\node [draw,fill=white, font=\large] at (rel axis cs: 0.53,-0.9){
  \shortstack[l]{
    \hspace*{18mm}\textbf{Method} \\
    \hspace{0.7em}\ref*{No_Compression_fake_legend_ade20k} \textsf{No compression} \cite{segdeformer} \\  
    \hspace{0.7em}\ref*{ours_fake_legend} \textsf{Distributed JD (\textbf{Ours})} \\
    \hspace{0.7em}\ref*{Ahuja_fake_legend_ade20k} \textsf{Distributed Baseline} 
  }
};

\node [draw,fill=white,font=\large] at (rel axis cs: 0.4,-1.35){\shortstack[l]{
 \textbf{FLOPs (G)}   \\
\strut \ref*{Ahuja_ablation_mitb0_ade20k} 434  \\
\strut \ref*{ours_mitb0_ade20k} \hspace{-0.35em} 7 \\
\strut \ref*{ours_mitb0_48_ade20k} \hspace{-0.3em} 5 \\
\strut \ref*{ours_mitb0_32_ade20k} \hspace{0.03em}3 
}};
\node [draw,fill=white,font=\large] at (rel axis cs: 0.702,-1.35){\shortstack[l]{
\hspace{0.05em} \textbf{$\#$ Params (M)}   \\
\strut \ref*{Ahuja_ablation_mitb0_ade20k}  1.36, $d=256$ \\ 
\strut \ref*{ours_mitb0_ade20k} 0.11, $d=64$ \\ 
\strut \ref*{ours_mitb0_48_ade20k} 0.06, $d=48$ \\  
\strut \ref*{ours_mitb0_32_ade20k} 0.03, $d=32$ 
}};

\end{tikzpicture}}
  \caption{\textbf{Ablation study on the internal dimension $d$} of our proposed {\normalfont \textsf{distributed JD}} vs.\ the transformer-based {\normalfont \textsf{distributed baseline}} with \textbf{\texttt{MiT-B5}} encoder: Computational complexity per image and model size on \textbf{ADE20K} $\mathcal{D}_{\mathrm{ADE20K}}^{\mathrm{val}}$ dataset. Here, the {\normalfont \textsf{distributed baseline}} employs $d=256$ \cite{segdeformer}.}
  \label{fig:ablation_comparison_ade20k}
\end{figure}
\subsection{ABLATION STUDY}
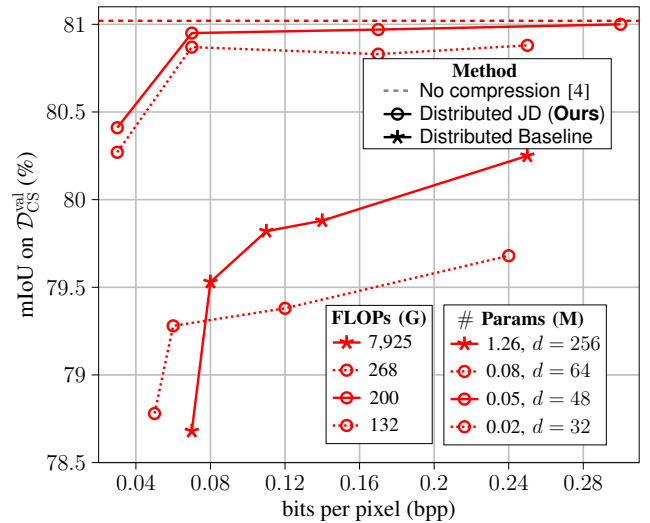
\begin{figure}[t!]
 \hspace{-0.4em}
    \resizebox{1\linewidth}{!}{\usetikzlibrary{shapes.geometric, arrows}
\usetikzlibrary{shapes.geometric, arrows.meta, calc, backgrounds}


\begin{tikzpicture}
\begin{axis}[
width=13.25cm,   
legend cell align={left},
legend cell align={left},
legend cell align={left},
legend style={
  fill opacity=1.0,
  draw opacity=1,
  text opacity=1,
  at={(0.5,0.02)},
  anchor=south west,
  draw=white!80!black
},
legend style={
  fill opacity=1.0,
  draw opacity=1,
  text opacity=1,
  at={(0.7,0.02)},
  anchor=south west,
  draw=white!80!black
},
legend style={
  fill opacity=1.0,
  draw opacity=1,
  text opacity=1,
  at={(0.7,0.02)},
  anchor=south west,
  draw=white!80!black
},
tick align=outside,
tick pos=left,
tick label style={font=\Large},
label style={font=\Large},
xlabel={bits per pixel (bpp) },
xmajorgrids,
xmin=0.02, xmax=0.31,
xtick distance=0.04,
xtick style={color=black},
x tick label style={/pgf/number format/precision=10},
x grid style={gray!50, line width=1pt},
y grid style={gray!50, line width=1pt},
ylabel={\(\displaystyle \textrm{mIoU on $\mathcal{D}_{\mathrm{CS}}^{\mathrm{val}}$}\) (\%)},
ylabel style={yshift=10pt},
xlabel style={yshift=-5pt},
ymajorgrids,
ymin=78.5, ymax=81.1,
ytick style={color=black},
ytick={78.5,79,79.5,80,80.5,81},
yticklabels={
\(\displaystyle {78.5}\),
\(\displaystyle {79}\),
\(\displaystyle {79.5}\),
\(\displaystyle {80}\),
\(\displaystyle {80.5}\),
\(\displaystyle {81}\),
 }
]
\addplot [line width=1.5pt,  red, dashed]
table {%
0.02 81.02
0.31 81.02
};
\label{No_Compression_ablation_mitb5_cs}

\addplot [line width=1.5pt,red,mark=star, mark size=4.5,  mark options={ line width=1.7, solid}]
table {%
0.07 78.68 
0.08 79.53
0.11 79.82
0.14 79.88
0.25 80.25
};
\label{Ahuja_ablation_mitb5_cs}

\addplot [line width=1.5pt,  red, dotted, mark=o, mark size=3.35,  mark options={ line width=1.5, solid}]
table {%
0.03 80.27
0.07 80.87
0.17 80.83
0.25 80.88
};
\label{ours_mitb5_32_cs}
\addplot [line width=1.5pt, dotted, red,  mark=o, mark size=3.35,  mark options={ line width=1.5, solid}]
table {
0.05 78.78
0.06 79.28
0.12 79.38
0.24 79.68
};
\label{ours_mitb5_64_cs}
\addplot [line width=1.5pt,  red,  mark=o, mark size=3.35,  mark options={ line width=1.5, solid}]
table {%
0.03 80.41
0.07 80.95
0.17 80.97
0.3 81
};
\label{ours_mitb5_48_cs}

\addplot [line width=1.5pt, gray , dashed]
table {%
0.0 78
};
\label{No_Compression_fake_legend_ade20k}
\addplot [line width=1.5pt,black,mark=star, mark size=4.5,  mark options={ line width=1.7, solid}]
table {%
0.0320000457763672 78
};
\label{Ahuja_fake_legend_ade20k}
\addplot [line width=1.5pt,black,mark=o, mark size=3.35,  mark options={ line width=1.5, solid}]
table {%
0.0320000457763672 78
};
\label{ours_fake_legend}
\end{axis}

\node [draw,fill=white,font=\large] at (rel axis cs: 0.65,0.595){ \shortstack[l]{
    \hspace*{18mm}\textbf{Method} \\
    \hspace{0.7em}\ref*{No_Compression_fake_legend_ade20k} \textsf{No compression} \cite{segdeformer} \\  
    \hspace{0.7em}\ref*{ours_fake_legend} \textsf{Distributed JD (\textbf{Ours})} \\
    \hspace{0.7em}\ref*{Ahuja_fake_legend_ade20k} \textsf{Distributed Baseline} 
  }
 };

\node [draw,fill=white,font=\large] at (rel axis cs: 0.45,0.001){\shortstack[l]{
 \textbf{FLOPs (G)}   \\
\strut \ref*{Ahuja_ablation_mitb5_cs} 7,925  \\
\strut \ref*{ours_mitb5_64_cs} \hspace{-0.35em} 268 \\
\strut \ref*{ours_mitb5_48_cs} \hspace{-0.3em} 200 \\
\strut \ref*{ours_mitb5_32_cs} \hspace{-0.3em} 132
}};
\node [draw,fill=white,font=\large] at (rel axis cs: 0.72,0.0001){\shortstack[l]{
\hspace{0.05em} \textbf{$\#$ Params (M)}   \\
\strut \ref*{Ahuja_ablation_mitb5_cs}  1.26, $d=256$ \\
\strut \ref*{ours_mitb5_64_cs} 0.08, $d=64$ \\ 
\strut \ref*{ours_mitb5_48_cs} 0.05, $d=48$ \\
\strut \ref*{ours_mitb5_32_cs} 0.02, $d=32$ 
}};
\end{tikzpicture}}
  \caption{\textbf{Ablation study on the internal dimension $d$} of our proposed {\normalfont \textsf{ distributed JD}} vs.\ the transformer-based {\normalfont \textsf{distributed baseline}} with \textbf{\texttt{MiT-B5}} encoder: Computational complexity per image and model size on \textbf{Cityscapes} $\mathcal{D}_{\mathrm{CS}}^{\mathrm{val}}$ dataset. Here, the {\normalfont \textsf{distributed baseline}} employs $d=256$ \cite{segdeformer}.}
  \label{fig:ablation_comparison_cs}
\end{figure}

In Figures \ref{fig:ablation_comparison_ade20k} and \ref{fig:ablation_comparison_cs}, we present ablation studies to select the internal dimension $d$ of our proposed transformer-based joint feature and task decoder $\mathbf{JD}$ (Figs.\ \ref{fig:overview_architectures}(b), \ref{fig:context_mining}, and \ref{fig:naive}) on the ADE20K $\mathcal{D}_{\mathrm{ADE20K}}^{\mathrm{val}}$ and Cityscapes $\mathcal{D}_{\mathrm{CS}}^{\mathrm{val}}$ datasets with an \texttt{MiT-B5} \cite{Segformer} encoder. We ablate over $d \in \{32, 48, 64\}$ and compare against the \textsf{distributed baseline}, which employs $d=256$ for both datasets \cite{segdeformer}. By varying $d$, we achieve different computational complexities per image and model size for $\mathbf{JD}$. Note that the investigated values of $d$ are applied consistently across all encoder variants, including \texttt{MiT-B0}, \texttt{MiT-B1} and \texttt{MiT-B2}, for both ADE20K and Cityscapes. As shown in Figures \ref{fig:ablation_comparison_ade20k} and \ref{fig:ablation_comparison_cs}, the configuration with $d=48$ produces the best RD performance consistently on $\mathcal{D}_{\mathrm{ADE20K}}^{\mathrm{val}}$ and on $\mathcal{D}_{\mathrm{CS}}^{\mathrm{val}}$, respectively. Accordingly, our proposed $\mathbf{JD}$ employs $d=48$. 

\section{CONCLUSIONS}

In this work, we demonstrate that the computational complexity of \texttt{SegDeformer} can be significantly reduced for both in-car and distributed applications through our proposed joint feature and task decoding approach. This largely reduces the required computational complexity for the in-car application and improves the scalability of the distributed semantic segmentation service, without putting a high computational burden onto the cloud. We show that our proposed approach increases the frames per second by up to a factor of $11.7$ (from $1.4$ fps to $16.5$ fps) on Cityscapes and by up to a factor of $3.5$ ($43.3$ fps to $154.3$ fps) on ADE20K for the in-car application, all while maintaining comparable mIoU to the transformer-based \textsf{no compression baseline}. Further, for the distributed application, we achieve state-of-the-art (SOTA) performance over a wide range of bitrates on the mIoU metric, while using only $0.14$\% ($0.04$\%) of cloud DNN parameters used in previous SOTA on ADE20K (Cityscapes) datasets.

\end{document}